\pdfoutput=1

\documentclass[11pt]{article}

\usepackage{acl}

\usepackage{times}
\usepackage{latexsym}
\usepackage{multirow}
\usepackage{makecell}
\usepackage{tabularx}
\usepackage{subcaption}
\usepackage{tablefootnote}

\usepackage[T1]{fontenc}

\usepackage[utf8]{inputenc}

\usepackage{microtype}

\usepackage{microtype}
\usepackage[normalem]{ulem}
\usepackage{times}
\usepackage{latexsym}
\usepackage{booktabs}
\usepackage[colorinlistoftodos]{todonotes}
\usepackage{enumitem}
\usepackage{xspace}
\usepackage{color, colortbl}
\usepackage{amssymb}
\usepackage{pifont}
\usepackage{algorithm,algpseudocode}
\usepackage{amsmath}
\usepackage{amsfonts}
\usepackage{amssymb}
\usepackage{listings}

\newcommand{\secref}[1]{\S\ref{#1}}
\newcommand{\benchmark}{{\fontfamily{lmtt}\selectfont CLUES}\xspace}
\newcommand{\benchmarkreal}{{\fontfamily{lmtt}\selectfont \benchmark-Real}\xspace}
\newcommand{\benchmarksyn}{{\fontfamily{lmtt}\selectfont \benchmark-Synthetic}\xspace}
\newcommand{\model}{{\fontfamily{lmtt}\selectfont ExEnt}\xspace}
\newcommand{\Thead}[1]{\textbf{\textsc{#1}}}
\definecolor{LightCyan}{RGB}{172,204,186}

\title{\benchmark: A Benchmark for Learning Classifiers using Natural Language Explanations}

\author{Rakesh R Menon\thanks{\hspace{0.5em}Equal contribution} \and Sayan Ghosh\footnotemark[1]  \and Shashank Srivastava\\
   UNC Chapel Hill \\ 
  \texttt{\{rrmenon, sayghosh, ssrivastava\}@cs.unc.edu}}

\begin{document}
\maketitle

\begin{abstract}
Supervised learning has traditionally focused on inductive learning by observing labeled examples of a task. In contrast, humans have the ability to learn new concepts from language. 
Here, we explore learning zero-shot classifiers for structured data\footnote{By structured data, we refer to data that can be reasonably represented using tables. This is a highly flexible format for representing a lot of real-world data (e.g., spreadsheets, traditional classification datasets in CSV format, single-table databases, as well as structured text-rich data such as emails), with a large variety in possible table schemas.} \emph{purely} from language from natural language explanations as supervision.
For this, we introduce \benchmark, a benchmark for \textbf{C}lassifier \textbf{L}earning \textbf{U}sing natural language \textbf{E}xplanation\textbf{S}, consisting of a range of classification tasks over structured data along with natural language supervision in the form of explanations. \benchmark consists of 36 real-world and 144 synthetic classification tasks. It contains crowdsourced explanations describing real-world tasks from multiple teachers and programmatically generated explanations for the synthetic tasks.
We also introduce \model, an entailment-based method for training classifiers from language explanations, which explicitly models the influence of individual explanations in making a prediction. \model generalizes up to 18\% better (relative) on novel tasks than a baseline that does not use explanations. We identify key challenges in learning from explanations, addressing which can lead to progress on \benchmark in the future. Our code and datasets are available at: \url{https://clues-benchmark.github.io}.
\end{abstract}

\section{Introduction}
\begin{figure}
    \centering
    \includegraphics[scale=0.6]{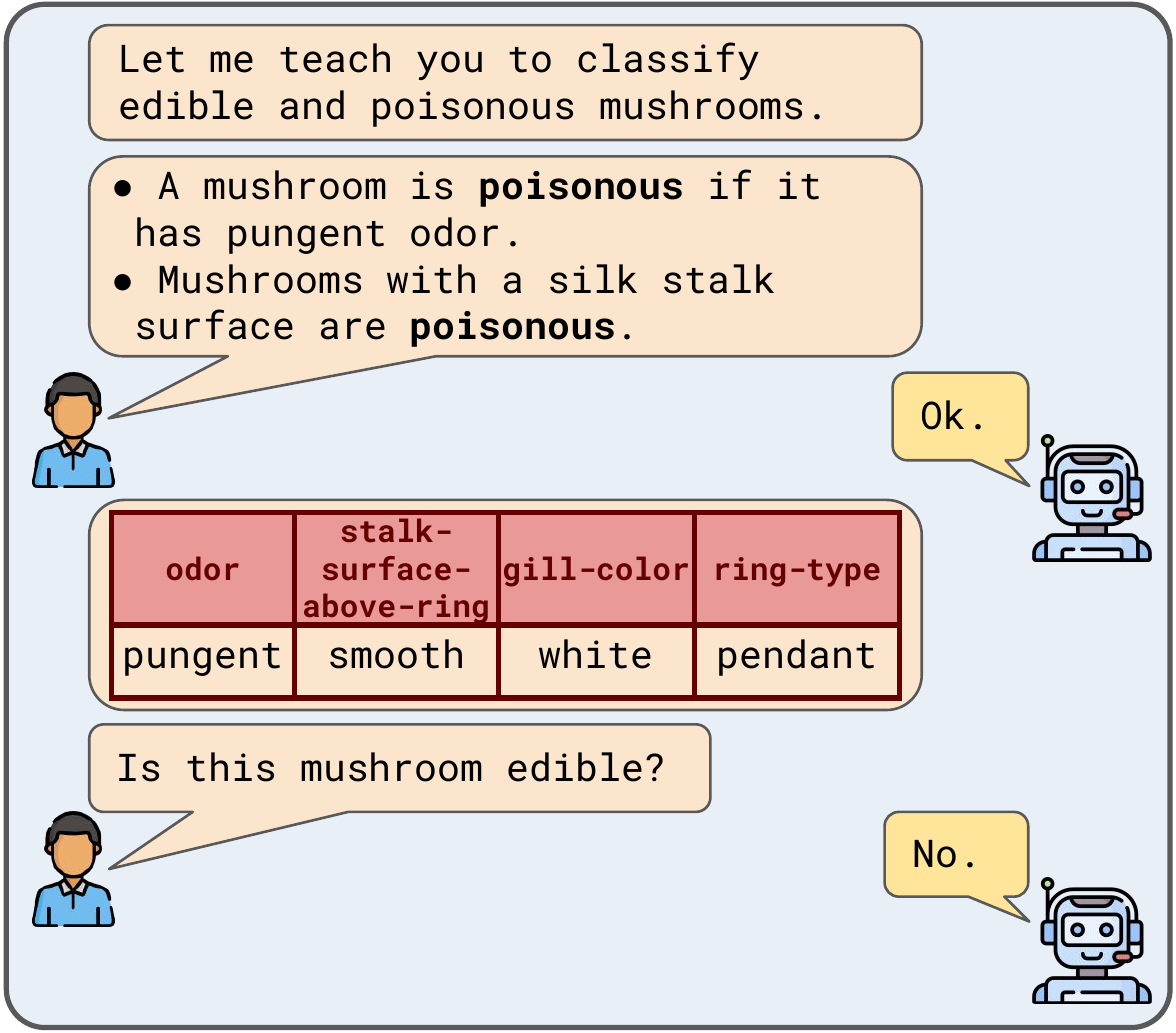}
    \caption{We explore learning classification tasks over structured data from natural language supervision in form of explanations. The explanations provide declarative supervision about the task, and are not example-specific. This is an example from the UCI Mushroom dataset, one of the 36 real-world datasets for which we collect multiple sets of explanations in \benchmark. 
    }
    \label{fig:intro}
    \vspace{-0.5cm}
\end{figure}

Humans have a remarkable ability to learn concepts through language \cite{chopra2019first, Tomasello1999TheCO}. For example, we can learn about \textit{poisonous mushrooms} through an explanation like \textit{`a mushroom is poisonous if it has pungent odor’}. Such an approach profoundly contrasts with the predominant paradigm of machine learning, where algorithms extract patterns by looking at scores of labeled examples of poisonous and edible mushrooms. 
However, it is unnatural to presume the availability of labeled examples for the heavy tail of naturally occurring concepts in the world. 

This work studies how models trained to learn from natural language explanations can generalize to novel tasks without access to labeled examples. While prior works in this area \cite{srivastava2017joint, srivastava-etal-2018-zero, hancock-etal-2018-training, murty-etal-2020-expbert, andreas-etal-2018-learning, Wang*2020Learning, ye2020teaching, zhou2020towards} 
have explored explanations as a source of supervision, they evaluate models on a small number of tasks (2-3 relation extraction tasks in \cite{hancock-etal-2018-training, Wang*2020Learning, murty-etal-2020-expbert, zhou2020towards}, 7 email categorization tasks \cite{srivastava2017joint}). 
Owing to the paucity of large-scale benchmarks for learning from explanations over diverse tasks, we develop \benchmark, a benchmark of classification tasks paired with natural language explanations.  Over the last few decades, researchers and engineers alike have put immense effort into constructing structured and semi-structured knowledge bases (e.g., structured tables on Wikipedia, e-commerce sites, etc.). Developing models that can reason over structured data is imperative to improve the accessibility of machine learning models, enabling even non-experts to interact with such data. Hence, in this work, we specifically formulate our classification tasks over structured data. 

Our benchmark is divided into \benchmarkreal and \benchmarksyn consisting of tasks from real-world (UCI, Kaggle, and Wikipedia) and synthetic domains respectively. Explanations for \benchmarkreal are crowdsourced to mimic the diversity and difficulty of human learning and pedagogy. 
For \benchmarksyn, we generate the explanations programmatically to explicitly test models’ reasoning ability under a range of structural and linguistic modifications of explanations.

We train models with a mix of explanations and labeled examples, in a multi-task setup, over a set of \emph{seen} classification tasks to induce generalization to \emph{novel} tasks, where we do not have any labeled examples.  
\newcite{ye-etal-2021-crossfit} refer to this problem setup as ``cross-task generalization". Some recent methods on cross-task generalization from language use instructions/prompts~\cite{Mishra2021CrossTaskGV, sanh2021multitask, wei2021finetuned} describing information about \textit{`what is the task?'} to query large language models. In contrast, language explanations in \benchmark provide the logic for performing the classification task, or intuitively \textit{`how to solve the task?'}. For the running example of mushroom classification, an instruction/prompt 
might be \textit{`can you classify a mushroom with pungent odor as poisonous or edible?'}. On the other hand, an example of an explanation in \benchmark is \textit{`a mushroom is poisonous if it has pungent odor'}.


We find that simply concatenating explanations to the input does not help pre-trained models, like RoBERTa \cite{liu2019roberta}, generalize to new tasks. Thus,  
we develop \model, an entailment-based model for learning classifiers guided by explanations, which explicitly models the influence of individual explanations in deciding the label of an example. 
\model shows a relative improvement of up to 18\% over other baselines on unseen tasks.

To identify the challenges of learning from explanations, we perform extensive analysis over synthetic tasks. Our analysis explores how the structure of an explanation (simple clauses vs. nested clauses) and the presence of different linguistic components in explanation (conjunctions, disjunctions, and quantifiers) affect the generalization ability of models.

The rest of the paper is structured as follows: we describe our crowdsourced-benchmark creation pipeline in \secref{sec:dataset_collection}. In \secref{sec:dataset_analysis}, we analyze our collected data. In \secref{sec:task_and_models}, we describe our models, experiments, and results.
We conclude with a brief discussion on the contributions and our findings, followed by a statement of ethics and broader impact.
Our contributions are:
\begin{itemize}[noitemsep, topsep=0pt, leftmargin=*]
    \item We introduce \benchmark, a benchmark for learning classifiers over structured data from language.
    \item We develop \model, an entailment-based model for learning classifiers guided by explanations.
    \model shows a relative improvement of up to 18\% over other baselines on generalization to novel tasks.
    \item We explore the effect on the generalization ability of models learning from language by ablating the linguistic components and structure of explanations over our benchmark's synthetic tasks. 
\end{itemize}


\section{Related Work}

\paragraph{Learning concepts from auxiliary information:} 
Prior work has explored techniques to incorporate `side-information' to guide models during training \cite{mann2010generalized, ganchev2010posterior}. More recently, researchers have explored using language in limited data settings for learning tasks such as text classification \cite{srivastava2017joint, srivastava-etal-2018-zero, hancock-etal-2018-training} and question answering \cite{Wang*2020Learning, ye2020teaching}. However, we diverge from these works by exploring the generalization ability of classifiers learned by using language over novel tasks as opposed to gauging performance only on seen tasks.

\noindent
\textbf{Explanation-based Datasets:} The role of explanations and how they can influence model behavior is a widely studied topic in machine learning \cite{wiegreffe-marasovic-2021-review}. Among language-based explanation studies, past work has primarily developed datasets that justify individual predictions made by a model (also called, local explanations) \cite{rajani2019explain, NEURIPS2018_4c7a167b}, \emph{inter alia}. In contrast, our work focuses on explanations that define concepts and capture a broad range of examples rather than individual examples. Our notion of explanations is shared with \citet{andreas-etal-2018-learning, srivastava2017joint, srivastava-etal-2018-zero}.
We differ from these works as (1) our benchmark comprises a large set of classification tasks spanning diverse concepts for learning from explanations as opposed to working on a limited set of tasks in prior work and (2) our benchmark is domain agnostic in the source of classification tasks considered as long as we can represent the inputs of the task in a tabular (structured) format.

\noindent
\textbf{Few-shot \& Zero-shot learning:} Large pre-trained language models (LMs) \cite{devlin2018bert, liu2019roberta, raffel2020t5} have been shown to perform impressively well in few-shot settings \cite{NEURIPS2020_1457c0d6, lester-etal-2021-power}. Reformulating natural language tasks with patterns has been shown to boost few-shot learning ability for small language models as well \cite{schick-schutze-2021-just, tam-etal-2021-improving}. 
More recently, a few works have focused on evaluating the generalization of models to unseen tasks by using prompts and performing multi-task training~\cite{Mishra2021CrossTaskGV, ye-etal-2021-crossfit, sanh2021multitask, min2021metaicl, chen2021meta, aghajanyan-etal-2021-muppet} While the training and evaluation setup is similar, our work is significantly different from these works as (1) the explanations in our work provide rationales for making a classification decision as opposed to explaining a task using prompts, (2) we explore classification over structured data as opposed to free-form text by designing a model that can leverage explanations.

\section{Creating \benchmark}
\label{sec:dataset_collection}

In this section, we describe our benchmark creation process in detail. 
In \benchmark, we frame classification tasks over structured data represented in tabular format.
Based on the source of tables used to construct the classification tasks, we consider two splits of our benchmark, \benchmarkreal (real-world datasets) and \benchmarksyn (synthetic datasets).

\subsection{\benchmarkreal}
We first gather/create classification tasks from UCI, Kaggle, and Wikipedia tables, then collect explanations for each classification task.
\subsubsection{Collecting classification datasets}
\paragraph{Classification tasks from UCI and Kaggle.}
UCI ML repository\footnote{\url{https://archive.ics.uci.edu/ml/}} and Kaggle\footnote{\url{https://www.kaggle.com/datasets}} host numerous datasets for machine learning tasks.
For our benchmark, we pick out the tabular classification datasets. Then, we manually filter the available datasets to 
avoid ones with (a) many missing attributes and (b) complex attribute names that require extensive domain knowledge making them unsuitable for learning purely from language. 
\benchmarkreal contains 18 classification tasks from UCI and 7 from Kaggle (the details of tasks are in Appendix~\ref{sec:uci_kaggle_tasks_appendix}).

\noindent
\textbf{Mining tables from Wikipedia.}
Wikipedia is a rich, free source of information readily accessible on the web. Further, a lot of this information is stored in a structured format as tables. We explore creating additional classification tasks based on tables from Wikipedia, where each row in a table is assigned a category label.
However, only a small fraction of the tables might be suitable to frame a classification task for our benchmark.
Thus, we need to identify suitable tables by \textit{mining} a large collection of tables from Wikipedia (we use Wikipedia dump available on April 2021).
We formalize this mining-and-pruning process as a crowdsourcing task (on Amazon Mechanical Turk), where we present each turker with a batch of 200 tables 
and ask them to pick out suitable tables from that batch.  
For a table considered suitable by a turker, we further ask the turker to mention which column of the table should be considered as providing the classification labels.
We identified 11 classification tasks corresponding to 9 Wikipedia tables after mining around 10K Wikipedia tables (the details of tasks are provided in Appendix~\ref{sec:uci_kaggle_tasks_appendix}).

\subsubsection{Explanation Collection Pipeline}

Our explanation collection process consists of two stages – (1) teachers providing explanations after reviewing multiple labeled examples of the task, and (2) students verifying explanations and classifying new examples based on explanations for the tasks.

\noindent
\textbf{Collecting explanations}:  
\label{sec:collection}
We use the Amazon Mechanical Turk (AMT) platform to collect explanations for \benchmarkreal.
In each HIT, we provide turkers with a few labeled examples of a dummy task (each corresponding to a row in a table) and a set of good and bad explanations for the task to teach them about the expected nature of explanations. Next, we test them on a `qualification quiz' to gauge their understanding of good explanations. 

Upon qualification, the turker advances to the explanation collection phase of the HIT. At this stage, the turker is provided with 15-16 labeled examples of a task in \benchmarkreal and we ask them to write explanations describing the logic behind the classification for each class. Turkers are required to submit a minimum of two explanations ($\geq 5$ tokens each) for each task. 

Further, teachers can test their understanding by taking a validation quiz, where they make predictions over new unlabeled examples from the task. Based on their informed classification accuracy, teachers can optionally refine their explanations.

Finally, when turkers are content with their performance, they `freeze' the explanations and advance to the test-quiz where they are evaluated on a new set of unlabeled examples from the task (different from validation quiz).\footnote{For reference, we show snapshots of our annotation interface in Appendix \secref{sec:annotation_interface}.}
 We will refer to turkers who have provided responses at this stage as `teachers' since they provide explanations to `teach' models about different classification tasks. 
\noindent
\textbf{Verification of explanations}:  
\label{sec:verification}
After the explanation collection, we validate the utility of the sets of explanations for a task from each teacher by evaluating if they are useful they are for other humans in learning the task. For this, a second set of turkers\footnote{59 turkers participated in this stage.} is provided access to the collected explanations from a teacher for a task, but no labeled examples. These turkers are then asked to predict the labels of test examples from the held-out test set, solely based on the provided explanations.  

Additionally, we ask turkers in the verification stage to give a Likert rating (1-4 scale) on the usefulness of each explanation. 
Since the turkers in the verification stage perform the classification task using language explanations from a teacher, we refer to them as `students' for our setup. 

Thus, the tasks in \benchmarkreal contain explanations from multiple teachers and multiple students corresponding to a teacher. This provides rich information about variance in teacher and student performance indicating how amenable different tasks are for learning via language. We provide insights into the performance of teachers and students of our setup in \secref{sec:dataset_analysis}.

\begin{table}[!t]
\small
    \centering
    \begin{subtable}[t]{0.23\textwidth}
    \begin{tabular}[t]{lc}
        \toprule
        \rowcolor{LightCyan}
        \multicolumn{2}{c}{\Thead{\benchmarkreal}}\\
        \# Binary &  26\\
         \# Multiclass & 10 \\
        Avg. \# Expls./task & 9.6 \\
         Avg. \# teachers & 5.4\\
         Avg. \# Expls./teacher & 2.3\\
         \# students/teacher & 3\\
        Max. \# examples & 65K \\
        Min. \# examples & 5\\
        Median. \# examples & 442 \\
        Avg. \# features & 5.6 \\
        \bottomrule
    \end{tabular}
    \end{subtable}
    \hfill
    \begin{subtable}[t]{0.23\textwidth}
    \begin{tabular}[t]{lc}
        \toprule
        \rowcolor{LightCyan}
        \multicolumn{2}{c}{\Thead{\benchmarksyn}}\\
      \# Task types & 48 \\
         \# Binary & 94  \\
         \# Multiclass  & 50 \\
        Avg. \# Expls./task & 1.7 \\
        \# Examples/task & 1000 \\
        \# features/task & 5 \\
        \bottomrule
    \end{tabular}
    \end{subtable}
    \caption{Statistics of tasks in \benchmark}
    \vspace{-0.4cm}
    \label{tab:task_statistics}
\end{table}
\vspace{-0.1cm}
\subsection{\benchmarksyn}
\label{sec:benchmarksyn_creation}
The complexity and fuzziness of real-world concepts and the inherent linguistic complexity of crowdsourced explanations can often shroud the aspects of the task that make it challenging for models to learn from explanations. 
To evaluate models in controlled settings where such aspects are not conflated, we create \benchmarksyn, a set of programmatically created classification tasks with varying complexity of explanations (in terms of structure and presence of quantifiers, conjunctions, etc.) and concept definitions.

\noindent
We create tasks in \benchmarksyn by first selecting a table schema from a pre-defined set of schemas, then generating individual examples of the task by randomly choosing values (within a pre-defined range, obtained from schema) for each column of the table. Next, we assign labels to each example by using a set of `rules' for each task. 
In this context, a `rule' is a conditional statement (analogous to conditional explanations that we see for real-world tasks) used for labeling the examples.
We use the following types of rules that differ in structure and complexity ($c_i$ denotes $i^{th}$ clause and $l$ denotes a label):
\begin{itemize}[noitemsep, topsep=0pt, leftmargin=*]
    \item Simple: \texttt{IF $c_1$ THEN $l$}
    \item Conjunctive: \texttt{IF $c_1$ AND $c_2$ THEN $l$}
    \item Disjunctive: \texttt{IF $c_1$ OR $c_2$ THEN $l$}
    \item Nested disjunction over conjunction: \texttt{IF $c_1$ OR ($c_2$ AND $c_3$) THEN $l$}
    \item Nested conjunction over disjunction: \texttt{IF $c_1$ AND ($c_2$ OR $c_3$) THEN $l$}
    
    \item For each of the above, we include variants with negations (in clauses and/or labels): Some examples--\texttt{IF $c_1$ THEN NOT $l$}, \texttt{IF $c_1$ OR NOT $c_2$ THEN $l$}
\end{itemize}
\noindent
We also consider other linguistic variations of rules by inserting quantifiers (such as `always', `likely'). 
The synthetic explanations are template-generated based on the structure of the rules used in creating the task.
For brevity, we defer additional details on the use of quantifiers, label assignment using rules, and creation of synthetic explanations to Appendix \ref{sec:app_syn_details}.
Overall we have 48 different task types (based on the number of classes and rule variants) using which we synthetically create 144 classification tasks (each containing 1000 labeled examples).

\vspace{-0.1cm}
\section{Dataset analysis}
\label{sec:dataset_analysis}

\noindent
In this section, we describe the tasks and the collected explanations in \benchmark.



\begin{figure*}[!t]
    \centering
\minipage{0.298\textwidth}
  \includegraphics[width=\linewidth]{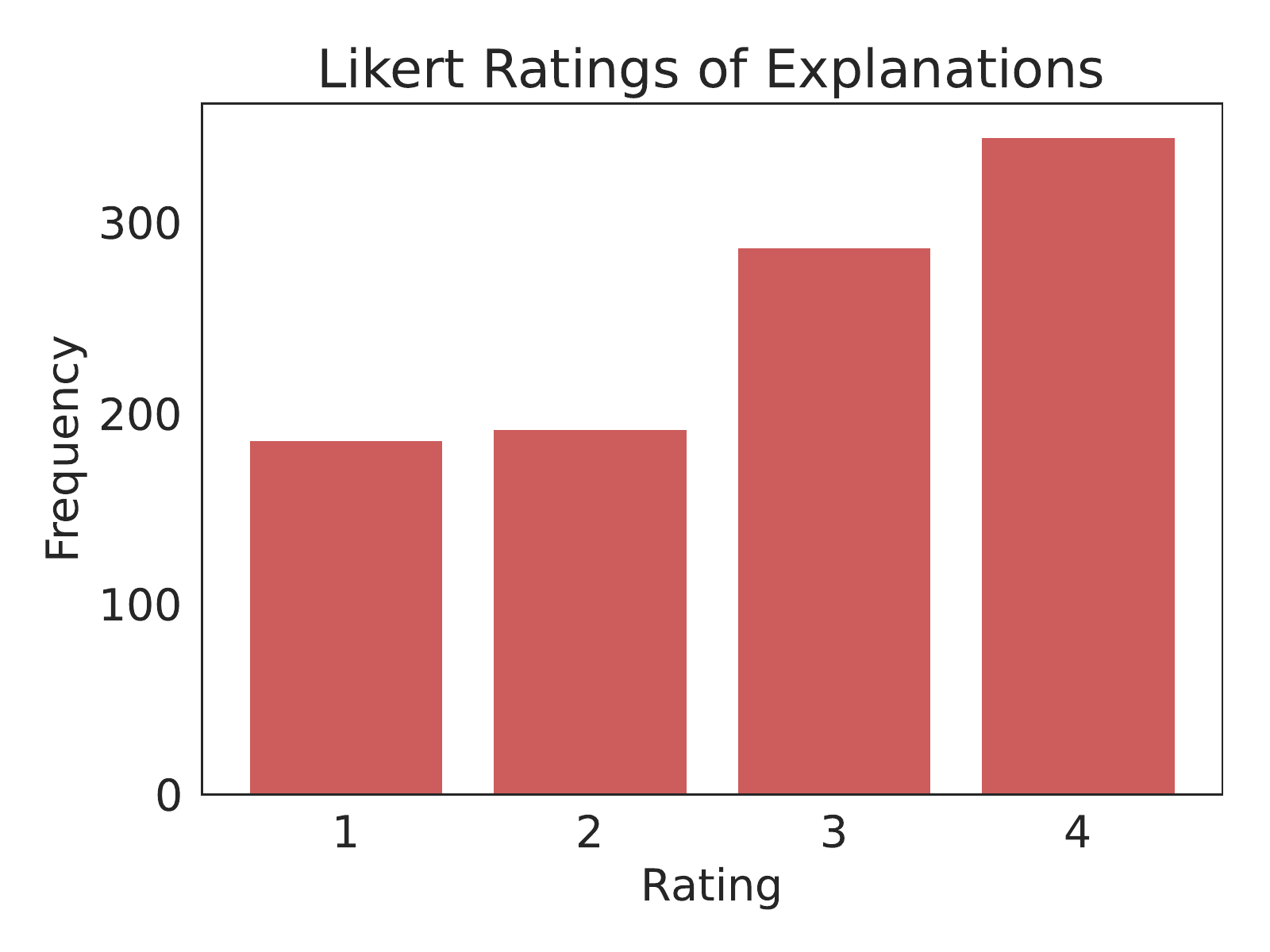}
  \centering
    \begingroup\renewcommand{\caption}[1]{(a)}%
  \caption{}
  \endgroup
\endminipage\hfill
\minipage{0.34\textwidth}
  \includegraphics[width=\linewidth]{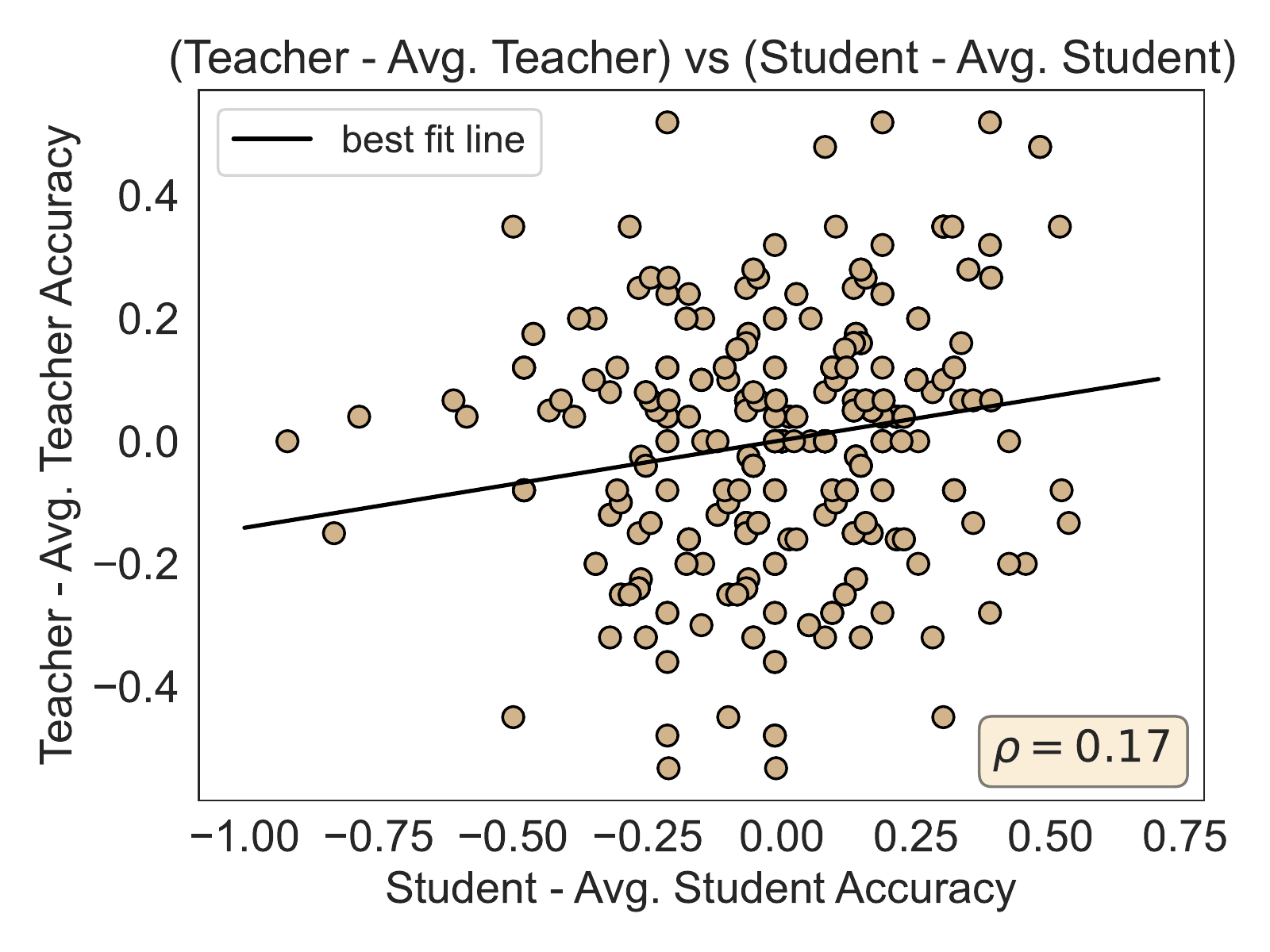}
  \centering
  \begingroup\renewcommand{\caption}[1]{(b)}%
  \caption{}
  \endgroup
\endminipage\hfill
\minipage{0.34\textwidth}%
  \includegraphics[width=\linewidth]{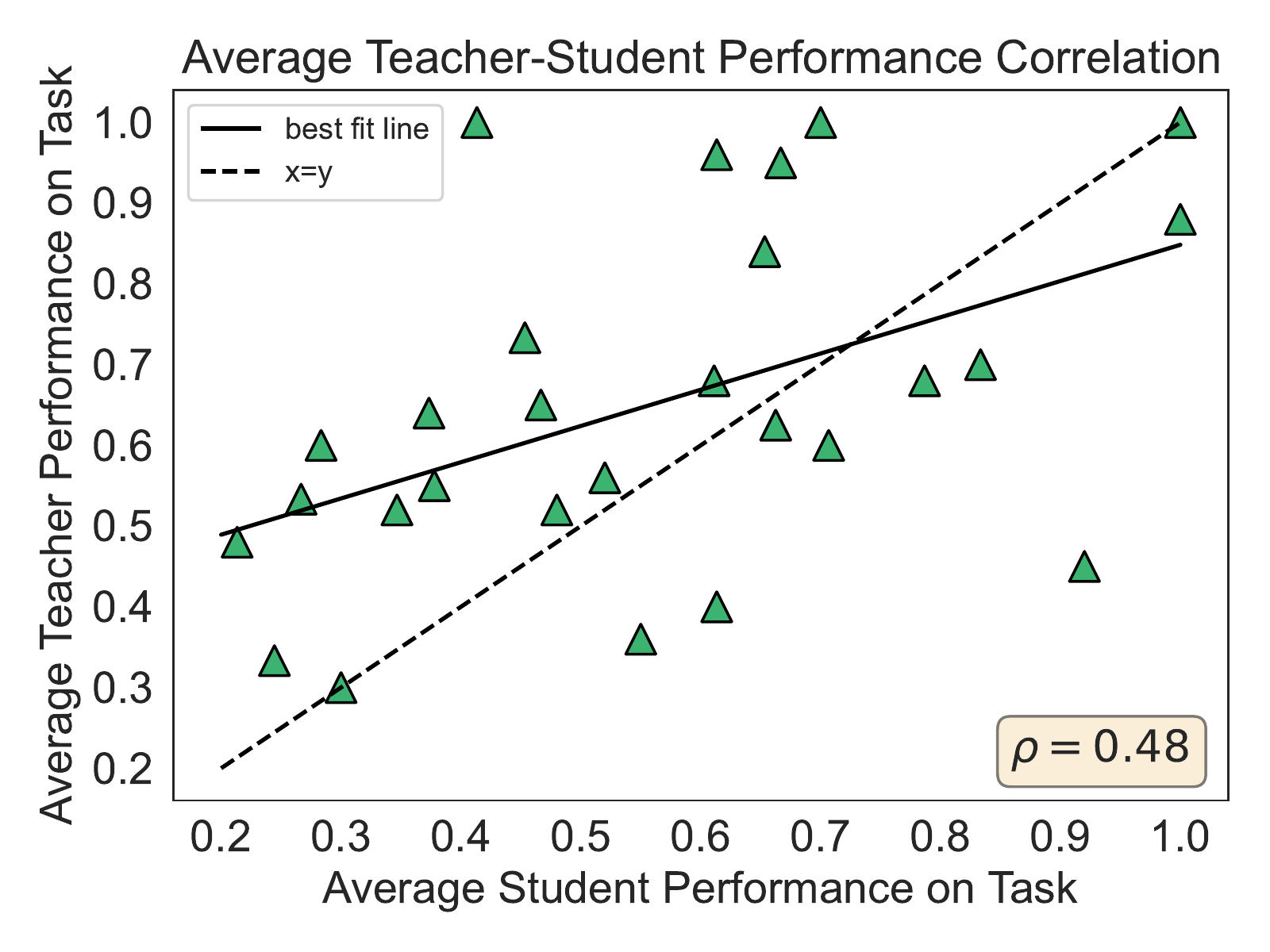}
\centering
  \begingroup\renewcommand{\caption}[1]{(c)}%
  \caption{}
  \endgroup
\endminipage
    \caption{(a) Histogram of count of explanations corresponding to different usefulness likert ratings. 
    (b) Students typically perform well when taught tasks by good teachers. (c) Positive correlation in the average performance between a teacher and student for a task. ($\rho$ denotes Pearson correlation coefficient in each of the plots)}
    \label{fig:variations_across_tasks}
    \vspace{-0.5cm}
\end{figure*}
\noindent
\textbf{Task Statistics}: 
Table~\ref{tab:task_statistics} shows the statistics of tasks in \benchmark. The real-world tasks in our benchmark are from a wide range of domains, such as data corresponding to a simple game (e.g. tic-tac-toe), medical datasets (e.g. identifying liver patients), merit-classification of teachers and students,  network-related datasets (eg. internet-firewall), among others. The synthetic tasks are created using table schemas denoting different domains, such as species of animals, species of birds, etc. (details in Appendix \ref{sec:app_syn_details}).

As seen in Table~\ref{tab:task_statistics}, 5.4 explanation sets were collected for each classification task from human teachers on average. Further, each explanation set was verified by 3 students during the verification task. An aggregate of 133 teachers provide 318 explanations for tasks in \benchmarkreal.  All collected explanations were manually filtered and irrelevant explanations were removed.
\begin{table}[!t]
\small
    \centering
    \begin{subtable}[t]{0.23\textwidth}
    \begin{tabular}{lc}
        \toprule
        Vocabulary & 1026\\
         Avg. \# tokens & 15.53\\
         Unique bigrams & 3300\\
        \bottomrule
    \end{tabular}
    \caption{Lexical statistics}
    \label{tab:dataset_explanation_statistics}
    \end{subtable}
    \hfill
    \begin{subtable}[t]{0.23\textwidth}
    \begin{tabular}{lc}
        \toprule
        Max. Score & 106.67 \\
        Min. Score &  3.12\\
        \bottomrule
    \end{tabular}
    \caption{Flesch Reading Complexity Scores}
    \label{tab:reading_complexity_statistics}
    \end{subtable}
    \caption{Explanations Statistics for \benchmark}
    \vspace{-0.4cm}
    \label{tab:overall_data_analysis}
\end{table}
\noindent
\textbf{Lexical analysis of explanations}: 
Table~\ref{tab:dataset_explanation_statistics} shows the statistics for explanation texts in our dataset.\footnote{Statistics in Table \ref{tab:dataset_explanation_statistics} was obtained using the \texttt{spacy} tokenizer.} We evaluate the average length of the explanation texts, vocabulary size and number of unique bigrams present in the explanations.

    

\begin{table}[!ht]
\small
    \centering
    \begin{tabular}{l|l|c|c}
    \toprule
    \Thead{Category} & \Thead{Example}
          & \Thead{Real} & \Thead{Syn} \\
    \midrule
         \textbf{Generic} & \makecell[l]{Being over 50 increases\\ the risk of a stroke.} & 48 \% & 50 \% \\[0.2cm]
         \textbf{Quantifier} & \makecell[l]{... \underline{\emph{usually}} means you\\ won't have heart disease.} & 52 \% & 50 \%\\[0.3cm]
         \textbf{Conditional} & \makecell[l]{\underline{\emph{If}} color code ... , \underline{\emph{then}} ...} & 15 \% & 100 \%\\[0.2cm]
        \textbf{Negations} & ... is \underline{\emph{not}} low. & 16 \% & 50\% \\
         \bottomrule
    \end{tabular}
    \caption{Count of explanations in our dataset based on various aspects of language present in them}
    \label{tab:explanation_categories}
    \vspace{-0.3cm}
\end{table}

\noindent
\textbf{Explanation characteristics}: 
Following \citet{chopra2019first}, we categorize the explanations based on the different aspects of language (generics, quantifiers, conditional, and negation) present in these explanations. Table~\ref{tab:explanation_categories} shows the statistics of various categories in our dataset. Note that an explanation might belong to more than one category (for example, an example like \textit{``if the number of hands equal to 2, then it is usually foo"}, will be categorized both as having both conditional and quantifiers.)
We found that around 52\% of the explanations for the real-world tasks had quantifiers (such as `some', `majority', `most', etc.) in them. 
A full list of quantifiers present in the data is given in Appendix~\ref{sec:app_syn_details}.

\begin{figure}[!ht]
    \includegraphics[scale=0.47]{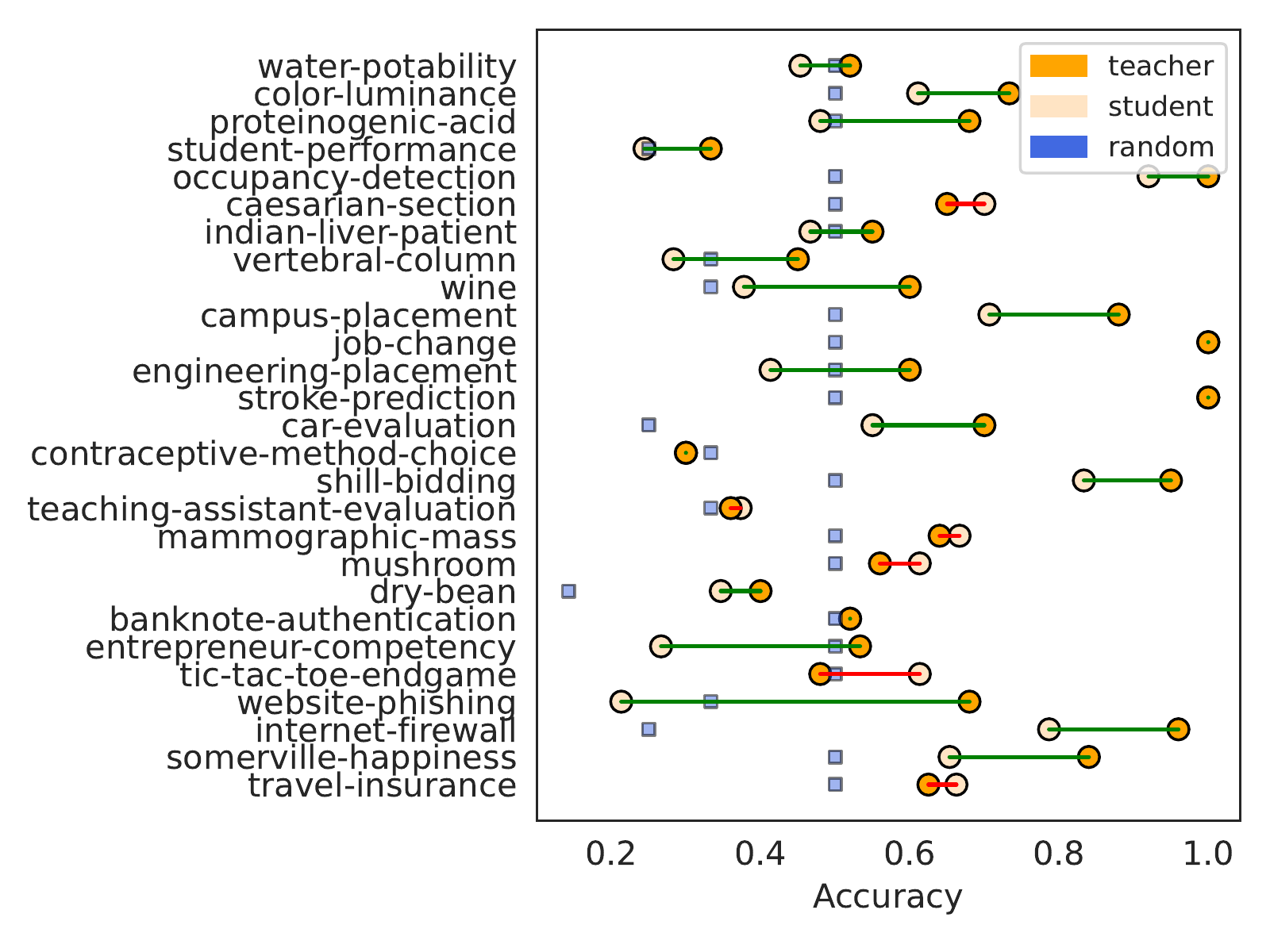}
    \caption{Average student vs average teacher performance for tasks in \benchmarkreal. 
    \textcolor{red}{Red} lines indicate cases where the student performance is more than the teacher performance. {\color{ForestGreen} Green} lines indicate cases where teachers perform better than students.}
    
    \label{fig:teacher_student}
    \vspace{-0.3cm}
\end{figure}

\noindent
\textbf{Reading complexity}: 
We analyze the reading complexity of crowdsourced explanations by using \textit{Flesch reading ease}\footnote{\url{https://en.wikipedia.org/wiki/Flesch_Kincaid_readability_tests}}.
Reading complexity values for our crowdsourced explanations vary from 3.12 (professional grade reading level) to 106.67 (easier than 3rd-grade reading level), with a median value of 65.73 (8th/9th-grade reading level). 
\noindent
\textbf{Usefulness of the explanations}: 
During the validation stage, we ask the turkers to provide a rating (on a Likert scale from 1 to 4) on the utility of the explanations for classification. The semantics of ratings are,  
1 -- `not helpful', 
2 -- `seems useful', 
3 -- `helped in predicting for 1 sample', and 
4 -- `mostly helpful in prediction'. The average rating for the explanations in \benchmarkreal is 2.78, denoting most explanations were useful, even if they did not directly help predict labels in some cases. In Figure \ref{fig:variations_across_tasks}(a), we also provide a histogram of the Likert ratings provided by the students.

\begin{table}[!ht]
\small
    \centering
    \begin{tabular}{l|c|c}
    \toprule
        \Thead{} & \Thead{Validation} & \Thead{Test}\\
        \midrule
        Teacher & 69\% & 64\%\\
        Student & - & 55\%\\
        \bottomrule
    \end{tabular}
    \caption{Teacher/student performance on \benchmarkreal
    }
    \label{tab:human_perf}
    \vspace{-0.4cm}
\end{table}
\noindent
\textbf{Characteristics of teachers and students}: 
Figure~\ref{fig:variations_across_tasks}(b) shows the normalized teacher performance vs normalized student performance for teacher-student pairs in \benchmarkreal. Normalized performance of an individual teacher (or, student) on a task is defined as the difference between the performances of the teacher (or, student) and an average teacher (or, student) for the same task.
The positive correlation ($\rho$ = 0.17) suggests that students tend to perform well if taught by well-performing teachers. 
Positive correlation ($\rho$ = 0.48) in Figure~\ref{fig:variations_across_tasks}(c), 
indicates that task difficulty (captured by classification accuracy) is well-correlated for a teacher and student on average.

On visualizing the difference between an average student and an average teacher performance for each task in \benchmarkreal,
we find that an average teacher performs better than the average student on most tasks. However, for the `tic-tac-toe' task in \benchmarkreal, we find that the student accuracy was around 13\% higher than average teacher performance. We hypothesize that this task can be solved by commonsense reasoning without relying on the provided explanations, resulting in students performing better than teachers.
We quantify the average performance of teachers and students on \benchmarkreal in Table~\ref{tab:human_perf}.\footnote{Note that teacher scores in the tables and figures do not include 9 Wikipedia Tasks for which the authors formed the explanations. These 9 datasets had extremely few samples ($\sim$5), so this procedure was adopted. The list of crowdsourced tasks can be found in Table \ref{tab:benchmark_source}.} We find that students perform lower than teachers on average as expected since a teacher has more expertise in the task. Moreover, it is challenging to teach a task perfectly using explanations in a non-interactive setting where a student cannot seek clarifications.

Additional data analysis and details of HIT compensation can be found in Appendix~\ref{sec:appendix_data_analysis} and \ref{sec:appendix_reward}.

\section{Experiment Setup and Models}
\label{sec:task_and_models}
In this section, we describe our training and evaluation setup, our models, and experimental findings.

\subsection{Training and Evaluation Setup}
\label{sec:training_setup}
\begin{figure}[!ht]
    \centering
    \includegraphics[width=0.48\textwidth]{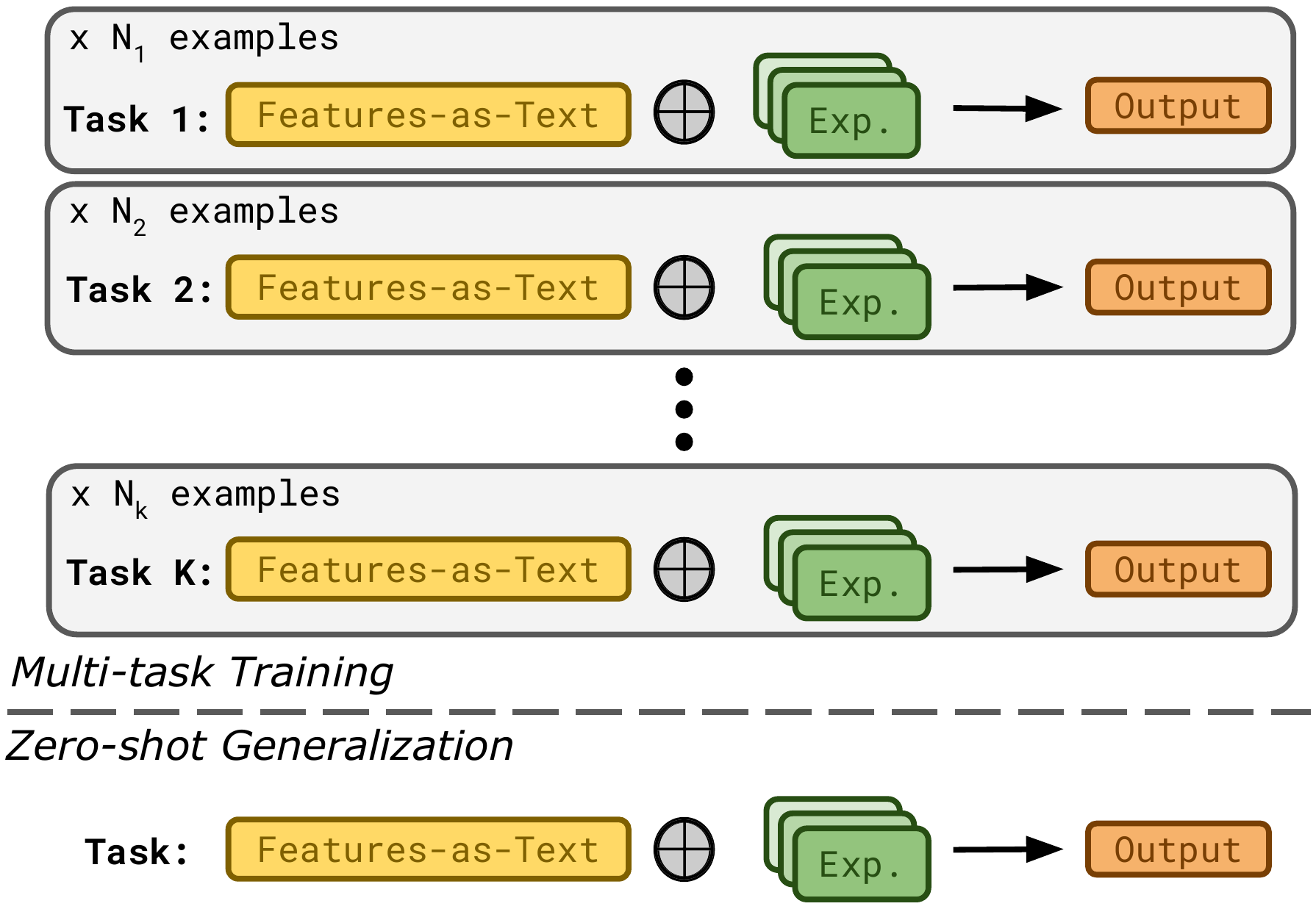}
    \caption{\textbf{Benchmark setup}: The model is trained on a set of classification tasks using explanations. At inference, the model is evaluated zero-shot on novel tasks using \textit{only} explanations for the novel tasks.}
    \label{fig:training_setup}
    \vspace{-0.3cm}
\end{figure}

Our goal is to learn a model that, at inference, can perform classification over an input $x$ to obtain the class label $y$, given the set of explanations $E$ for the classification task.
Figure~\ref{fig:training_setup} shows our setup, where we train our model using multi-task training over a set of tasks $\mathcal{T}_{seen}$ and evaluate generalization to a new task, $t \in \mathcal{T}_{novel}$. The task split we use for our experiments can be found in Appendix \ref{sec:task_split}. We select our best model for zero-shot evaluation based on the validation scores on the seen tasks. Since we do not make use of any data from the novel tasks to select our best model, we maintain the \emph{true zero-shot} setting \cite{perez2021true}.

\begin{figure*}[!t]
    \centering
    \includegraphics[scale=0.65]{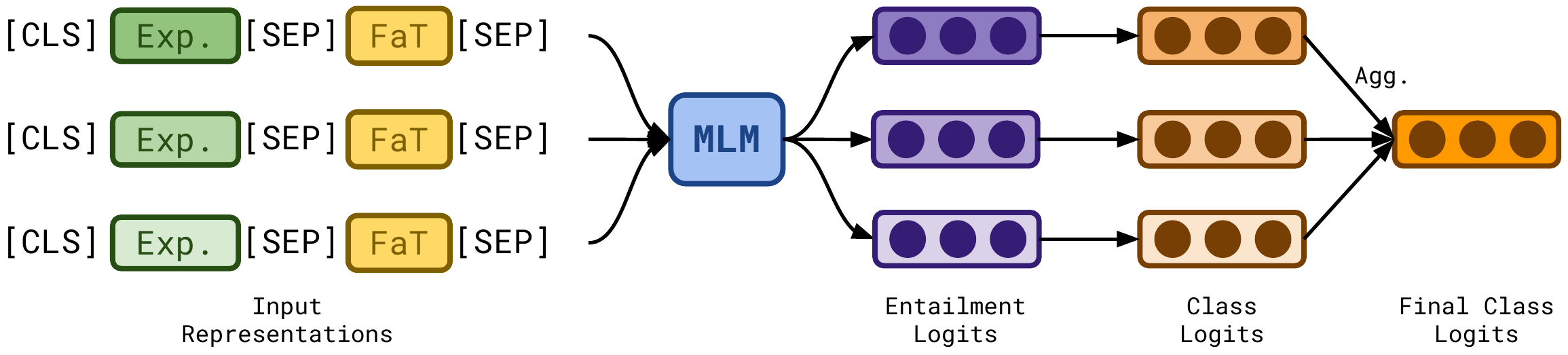}
    \caption{\model takes in concatenated pairs of individual task explanations and features of an example as input and uses a masked language model (MLM) to compute an entailment score for every explanation-feature pair of a task. Next, we map the entailment scores to class logits and finally apply an aggregation function over all the logits to obtain a final class prediction for the example.}
    \label{fig:proposed_model}
    \vspace{-0.5cm}
\end{figure*}

We encode each structured data example, $x$,  as a text sequence, by linearizing it as a sequence of attribute-name and attribute-value pairs, separated by \texttt{[SEP]} tokens. To explain, the leftmost attribute-name and attribute-value pair of structured input example in Figure~\ref{fig:intro} is represented as \texttt{`odor | pungent'}. The linearization allows us to make use of pre-trained language models for the classification task. 
Our linearization technique is similar to the one used in \citet{yin-etal-2020-tabert} with the exception that we do not use the column type. 
We will refer to the linearized format of structured inputs by `Features-as-Text' or `FaT'. 

\subsection{Baseline models}
\label{sec:baseline_models}

For our baselines, we make use of a pre-trained RoBERTa model \cite{liu2019roberta}.
However, RoBERTa with the standard-fine-tuning approach cannot allow a generalization test as the number of output classes varies for each task. 
Furthermore, we cannot train individual class heads at inference since we test \emph{zero-shot}. 
Hence, we make the following modifications to make RoBERTa amenable for zero-shot generalization tests: a pre-trained RoBERTa model takes the linearized structured data (FaT) as input and outputs a representation for this context (in the \texttt{[CLS]} token). Next, we run another forward pass using RoBERTa to obtain a representation of the labels based on their text (e.g., `poisonous' or `edible' for our example in Figure~\ref{fig:intro}). Finally, we compute the probability distribution over labels by doing a dot-product of the representations of the input and the labels. We train this model using cross-entropy loss. In our experiments, we refer to this model as RoBERTa w/o Exp since the model does not use any explanations.

We also experiment with a RoBERTa w/ Exp. model where a RoBERTa model takes as input a concatenated sequence of all the explanations for the task along with FaT. 
The rest of the training setup remains the same as RoBERTa w/o Exp.

\begin{figure}[t!]
    \centering
    \includegraphics[width=0.48\textwidth]{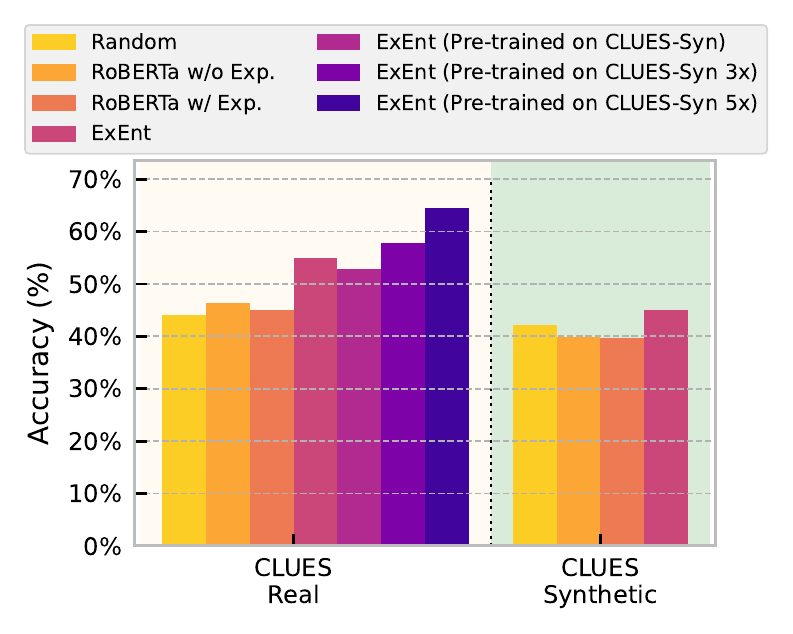}
    \caption{Zero-shot generalization performance of models on novel tasks of \benchmark.}
    \label{fig:zero_shot}
    \vspace{-0.6cm}
\end{figure}

We find that a simple concatenation of explanations is not helpful for zero-shot generalization to novel tasks (results in Figure~\ref{fig:zero_shot}).
Next, we describe \model which explicitly models the role of each explanation in predicting the label for an example.
\begin{figure*}[!ht]
    \centering
    \includegraphics[width=\textwidth]{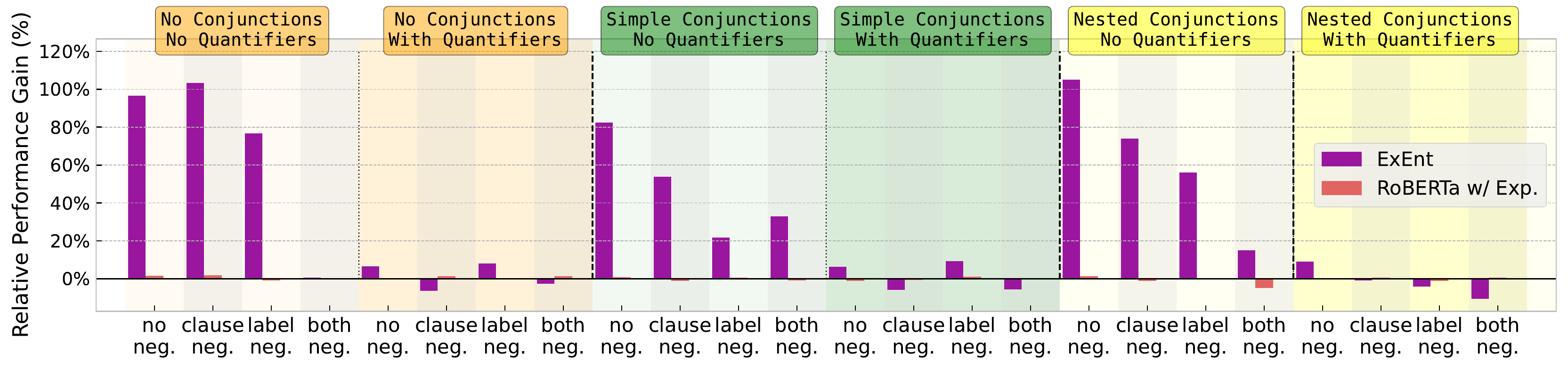}
    \caption{Ablation analysis on the effect of structural and linguistic variations of explanations on generalization ability of models. All bars indicate the relative performance gain over the RoBERTa w/o Exp. baseline.}
    \label{fig:ablation}
    \vspace{-0.5cm}
\end{figure*}

\subsection{\model}
To model the influence of an explanation towards deciding a class label, we draw analogies with the entailment of an explanation towards the structured input.
Here, given a structured input (\emph{premise}) and an explanation (\emph{hypothesis}), we need to decide whether the explanation strengthens the belief about a specific label (\emph{entailment}), weakens belief about a specific label (\emph{contradiction}) or provides no information about a label (\emph{neutral}).

Figure~\ref{fig:proposed_model} shows the overview of our explanation-guided classification model, \model; given a structured input and explanation of a task, 
let $l_{exp}$ denote the label mentioned in the explanation, and $L$ denote the set of labels of the task. 
The entailment model assigns logits $p_e$, $p_c$ and $p_n$ to the hypothesis being entailed, contradicted or neutral respectively w.r.t. the premise. Based on the label assignment referred to by an explanation, we assign logits to class labels as follows:  
\begin{itemize}[noitemsep, topsep=0pt, leftmargin=*]
    \item \textbf{If explanation mentions to assign a label} : 
    Assign $p_e$ to $l_{exp}$, $p_c$ is divided equally among labels in $L \setminus \{l_{exp}\}$, and $p_n$ is divided equally among labels in $L$. 
    \item \textbf{If explanation mentions to \underline{not} assign a label} : This occurs if a negation is associated with $l_{exp}$.
    Assign $p_c$ to $l_{exp}$, $p_e$ is divided equally among labels in $L \setminus \{l_{exp}\}$, and $p_n$ is divided equally among labels in $L$.
\end{itemize}
We obtain logit scores over labels of the task corresponding to each explanation as described above. We compute the final label logits by aggregating (using mean) over the label logits corresponding to each explanation of the task. The final label logits are converted to a probability distribution over labels, and we train \model using cross-entropy loss. 

In experiments, we consider a pre-trained RoBERTa model fine-tuned on MNLI~\cite{williams2017broad} corpus as our base entailment model.\footnote{Weights link: \url{https://huggingface.co/textattack/roberta-base-MNLI}} Further, in order to perform the assignment of logits using an explanation, we maintain meta-information for each explanation to (1) determine if the explanation mentions to `assign' a label or `not assign' a label, and (2) track $l_{exp}$ (label mentioned in explanation). For \benchmarksyn, we parse the templated explanations to obtain the meta-information, while for the explanations in \benchmarkreal, the authors manually annotate this meta-information.
Additional training details and hyperparameters are provided in Appendix~\ref{sec:model_appendix}.

\subsection{Zero-Shot Generalization Performance} \label{sec:experiments}
We evaluate \model and the baselines on zero-shot generalization to novel tasks in our benchmark as described in \secref{sec:training_setup}. 
We train separate models for \benchmarkreal and \benchmarksyn.
Figure~\ref{fig:zero_shot} shows the generalization performance of all models. On \benchmark, we find that \model outperforms the baselines 
suggesting that performing entailment as an intermediate step helps aggregate information from multiple explanations better. 
On \benchmarkreal, \model gets an 18\% relative improvement over the baselines while having an 11\% relative improvement on \benchmarksyn

To evaluate the utility of our synthetic tasks in enabling transfer learning to real-world tasks, we fine-tune a \model model pre-trained on synthetic tasks.
We experiment with three pre-training task sets - \benchmarksyn, \benchmarksyn (3x) and \benchmarksyn (5x) consisting of 144, 432, and 720 tasks.
These larger synthetic task sets are created by sampling tasks from each of the 48 different synthetic tasks types similar to how \benchmarksyn was created (see \secref{sec:benchmarksyn_creation} for reference).
We find that pre-training on synthetic tasks boosts the performance of \model on the novel tasks of \benchmarkreal by up to 39\% (relative) over the RoBERTa w/o Exp. model.

\paragraph{Human Performance}
To situate the performance of the automated models, we performed human evaluation for tasks in test split of \benchmarkreal using AMT. For this, we sampled at most 50 examples \footnote{Many tasks (such as tasks created from Wikipedia tables) have less than 50 examples in their test split.)} from the test split of tasks in \benchmarkreal and each example was `labeled' by 2 turkers using the explanations of the `best teacher' (the teacher whose students got the best performance during `explanation verification' stage; see \secref{sec:verification} for reference). The average human accuracy for this was about 70\%. However, the performance numbers of humans and models are not directly comparable as the model looks at all the explanations for the task, whereas the humans observe a small number of explanations. Humans also see multiple examples of the task during the evaluation, which they can use to fine-tune their understanding of a concept. The automated models don't have a mechanism to leverage such data.

\vspace{-0.1cm}

\section{Key Challenges} 
\label{sec:key_challenges}
To identify key challenges in learning from explanations, we perform experiments ablating the linguistic components
and structure of explanations.
For a robust analysis, we generate more tasks for each task type in \benchmarksyn, making 100 tasks for each of the 48 different task-types in \benchmarksyn (axes of variation include 4 negation types, 3 conjunction/disjunction types, 2 quantifier types, and number of labels; details in Appendix \ref{sec:synthetic_task_types}). 

We evaluate the generalization performance of \model to novel tasks on each of the different types separately by training separate models for each task type.
Figure~\ref{fig:ablation} shows the relative gain in generalization performance of models learned using explanations compared to the performance of baseline RoBERTa w/o Exp.\footnote{Accuracies have been averaged over the multi-class and binary datasets since the trends remain the same across both.}
Our results indicate that learning from explanations containing quantifiers is highly challenging.
In the presence of quantifiers, models guided by explanations perform on par with the baseline RoBERTa w/o Exp model.
Negations also pose a challenge, as indicated by the decline in relative gains of models guided by explanation compared to the RoBERTa w/o Exp model. 
Structurally complex explanations (containing conjunctions/disjunctions of clauses) are also hard to learn from compared to simple conditional statements. 
These challenges provide a fertile ground for future research and improvements.

\section{Conclusion}
We have introduced \benchmark, a benchmark with diverse classification tasks over structured data along with natural language explanations to learn them. \benchmark is agnostic in the domain of tasks allowing the research community to contribute more tasks in the future. We also present \model, an entailment-based model to learn classifiers guided by explanations. Our results are promising and indicate that explicitly modeling the role of each explanation through entailment can enable learning classifiers for new tasks from explanations alone. 
Future work can explore the open challenges in learning from explanations, such as modeling the influence of quantifiers and negations present in an explanation. 

Our empirical analyses here aggregates explanations for a task from multiple teachers. Future work can explore learning from explanations from individual teachers, as well as cross-teacher variance. 
Alternatively, rather than treat explanations from different teachers homogeneously, 
future work can model trustworthiness of a crowd of teachers from their provided explanations. 


\section*{Ethics and Broader Impact}
All tables in \benchmarkreal were collected from free public resources (with required attributions) and tables in \benchmarksyn were created by us programmatically. 
We do not collect any personal information from the turkers who participated in our crowdsourced tasks. The dataset has been released without mentioning any personal details of turkers available automatically in AMT (such as turker IDs). The turkers were compensated fairly and the payment per task is equivalent to an hourly compensation that is greater than minimum wage (based on the median time taken by turkers). We provide details of the reward structure for the crowdsourcing tasks in Appendix \ref{sec:appendix_reward}.
For the Wikipedia mining task in this work, we limited the locale of eligible turkers to US, UK, New Zealand and Australia. For other crowdsourcing tasks, we limited the locale of eligible turkers to US. Further, to ensure good-faith turkers, we required that the approval rate of the turkers be above 98\%. 
Our screening process has selection biases that likely over-samples turkers from demographics that are over-represented on AMT (ethnically white, college-educated, lower-to-medium income and young)~\cite{hitlin20164}, and this is likely to affect the type of language usage in the collected explanations. 

The broader impact of this research in the longer term could make developing predictive technologies more accessible to ordinary users, rather than data-scientists and experts alone. 

\nocite{strokeprediction, 9058292}

\bibliography{custom}
\bibliographystyle{acl_natbib}

\clearpage

\appendix
\section*{Appendix}
\begin{figure*}[!ht]
    \centering
    \includegraphics[width=\textwidth]{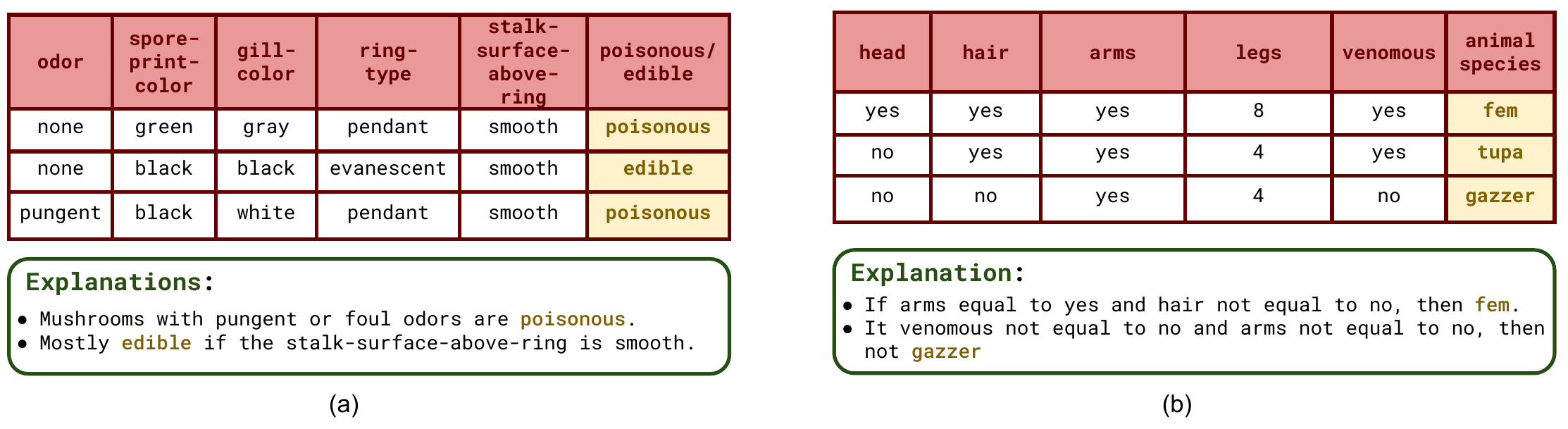}
    \caption{Example of tasks from \benchmark. The left and right tables are sample tables and explanations drawn from \benchmarkreal and \benchmarksyn respectively.}
    \label{fig:examples}
    \vspace{-0.3cm}
\end{figure*}


\section{Additional details on creating \benchmarksyn} \label{sec:app_syn_details}
In this section we discuss in detail about the various table schemas followed by the details of quantifiers and label assignment for creating synthetic tasks. 
\subsection{Tables schemas}
We define five different table schemas, each corresponding to a different domain.
For all the attributes in a schema we define a fixed domain from which values for that attribute can be sampled. 
\begin{itemize}[noitemsep, topsep=0pt, leftmargin=*]
    \item \textbf{Species of bird}: The classification task here is to classify a bird into a particular species based on various attributes (column names in table). We define several artificial species of birds using commonly used nonce words in psychological studies \cite{chopra2019first} such as ``dax", ``wug", etc. 
    
    \item \textbf{Species of animal}: The classification task here is to classify an animal into a particular species based on various attributes (column names in table). Artificial species of animals are again defined using commonly used nonce words in psychological studies such as ``dax", ``wug", etc. 
    
    \item \textbf{Rainfall prediction}: This is a binary classification task where the objective is to predict whether it will rain tomorrow based on attributes such as ``location", ``minimum temperature", ``humidity", ``atmospheric pressure" etc.
    
    \item \textbf{Rank in league}: This is a multi-label classification task where given attributes such ``win percentage", ``power rating", ``field goal rating" of a basketball club, the objective is to predict its position in the league out of {1, 2, 3, 4, "Not qualified"}.

    \item \textbf{Bond relevance}: This is a multi-label classification task where given attributes such ``user age", ``user knowledge", ``user income", the objective is to predict the relevance of a bond out of 5 classes (1 to 5).
\end{itemize}
In each of the above schemas, the attributes can be either of types categorical or numeral. 
For each of the above schemas we also define range of admissible values for each attribute. Detailed description of schemas are provided in Tables~ \ref{tab:schema_birds}, \ref{tab:schema_animals}, \ref{tab:schema_rain} \ref{tab:schema_team}, \ref{tab:schema_product}. 

\subsection{List of quantifiers}
The full list of quantifiers along with their associated probability values are shown in Table~\ref{tab:quantifier}.
\begin{table}[!ht]
\small
    \centering
    \begin{tabular}{l|c}
    \toprule
    \Thead{Quantifiers} & \Thead{Probability}\\
    \midrule
    "always", "certainly", "definitely" & 0.95 \\
    \makecell[l]{"usually", "normally", "generally",\\ "likely", "typically"} & 0.70  \\
    "often" & 0.50 \\
     "sometimes", "frequently", & 0.30\\
     "occasionally" & 0.20 \\
    "rarely", "seldom" & 0.10\\
     "never" & 0.05\\
     \bottomrule
    \end{tabular}
    \caption{Probability values used for quantifiers in \benchmarksyn. We choose these values based on \citet{srivastava-etal-2018-zero}.}
    \label{tab:quantifier}
\end{table}

\subsection{Creating synthetic explanations}
We use a template-based approach to convert the set to rules into language explanations. We convert every operator in the clauses into their corresponding language format as:
\begin{itemize}[noitemsep, topsep=0pt, leftmargin=*]
    \item $==$ \hspace{0.5cm}$\rightarrow$ `equal to'
    \item $ > $  \hspace{0.75cm}$\rightarrow$`greater than'
    \item $>= $ \hspace{0.5cm}$\rightarrow$`greater than or equal to'
    \item $< $ \hspace{0.75cm}$\rightarrow$`lesser than'
 \item $<= $ \hspace{0.5cm}$\rightarrow$`lesser than or equal to'
    \item $!=$ \hspace{0.5cm}$\rightarrow$ "not equal to'
    \item $!>$ \hspace{0.5cm}$\rightarrow$ `not greater than'
    \item $!< $ \hspace{0.5cm}$\rightarrow$ `not lesser than'
\end{itemize}
\noindent
For example if we have a rule `\texttt{IF number of hands == 2 THEN foo}', we convert it into a language explanation as `If number of hands equal to 2, then foo'.
In the presence of quantifiers, we add `it is \texttt{[INSERT QUANTIFIER]}' before the label. 
For example if the rule was associated with a quantifier `usually', the language explanation would be `If number of hands equal to 2, then it is usually foo'.

\subsection{Label Assignment using Rules}
In Algorithm \ref{alg:lab_assign}, we detail the procedure for obtaining label assignments for our synthetic tasks. Given that our rules are in an ``\texttt{IF} ... \texttt{THEN} .." format, we split each rule into an antecedent and a consequent based on the position of \texttt{THEN}. Note that our voting-based approach to choose the final label for an example helps to tackle (1) negation on a label for multiclass tasks and (2) choose the most suited label in case antecedents from multiple rules are satisfied by an example. 

\subsection{Different synthetic task types}
\label{sec:synthetic_task_types}
We create our synthetic tasks by varying along the following axes:
\begin{itemize}[noitemsep, topsep=0pt, leftmargin=*]
    \item Number of labels: $\mathbb{L} = $ \{`binary', `multiclass' \}
    \item Structure of explanation: $\mathbb{C} = $ \{`simple', `conjunction/disjunction', `nested' \}
    \item Quantifier presence: $\mathbb{Q}= $ \{`not present', `present'\}
    \item Negation: $\mathbb{N} = $ \{`no negation', `negation only in clause', `negation only on label', `negation in clause or on label'\}
\end{itemize}

The set of task types is defined as $\mathbb{L} \times \mathbb{C} \times \mathbb{Q} \times \mathbb{N}$, enumerating to 48 different types.

\subsection{Large synthetic task collections for ablation experiment}
In section~\secref{sec:key_challenges} we describe an ablation experiment, for which we create collections of 100 tasks corresponding to each synthetic task type. Here, the task type of a collection denotes the maximum complexity of explanations in that collection. For example, for the collection `multiclass classification with nested clauses and negation only in clause', all the 100 tasks might not have negations or nested clauses in their explanations. This collection might contain explanations with no negations or non-nested clauses. However, it will not contain explanations that have nested clauses and negations in both clause and label.

\begin{algorithm}[t!]
\small
	\caption{Label Assignment}
	\label{alg:lab_assign}
    \begin{algorithmic}[1]
    \State \textbf{Given:} Task $\mathcal{T}$ with rule set $R$ and label set $L$
    \State Votes $\gets$ Zeros($|L|$)
	\For{rule $r \in R$}
	    \State $r_a$ : Antecedent of $r$
	    \State $r_c$ : Consequent of $r$
		\State $l_{r} \gets $ Label mentioned in $r_c$
		\State $ t \gets $ Truth Value of $r_a$
		
		\If{any quantifier in $r$}
		    \State $p_{quant}$ : Prob. of quantifier from Table~\ref{tab:quantifier}
		    \State Alter $l_r$ to any label in $L \setminus {l_r}$ with probability \\\hspace{0.9cm} $1 - p_{quant}$
		\EndIf
		\If{$t = $ True}
		    \State Votes[$l_{r}$] += 1
		\Else
		    \For{label $l \in L \setminus {l_r} $}
		        \State Votes[$l$] += 1
		    \EndFor
		\EndIf
	    \State $l_{assigned} \gets$ argmax(Votes)
	\EndFor
    \end{algorithmic}
	\label{alg:training}
\end{algorithm} 

\section{Real-World Tasks from UCI, Kaggle and Wikipedia} 
\label{sec:uci_kaggle_tasks_appendix}

For our benchmark, we made use of 18 datasets in UCI, 7 datasets in Kaggle, and 9 tables in Wikipedia. In Table \ref{tab:benchmark_source}, we list the keywords that we use to refer to these tasks along with the URLs to the datasets/tables.

\subsection{Feature Selection for Real-World Datasets}

During pilot studies for collection of explanations for \benchmarkreal, we identified that annotators found it difficult to provide explanations for classifications tasks with more than 5 to 6 columns. Appropriately, we reduced the number of columns in most datasets of \benchmarkreal (apart from some Wikipedia tables) to 5 by choosing the top features that had maximum mutual information with the labels in the training dataset. The mutual information between the features and the label was computed using the \href{https://scikit-learn.org/stable/}{scikit-learn} package with a random state of 624.


\section{Additional Analysis on Teacher-Student Performance} \label{sec:appendix_data_analysis}

\begin{figure*}[t!]
   \centering
    \minipage{0.33\textwidth}
      \includegraphics[width=\linewidth]{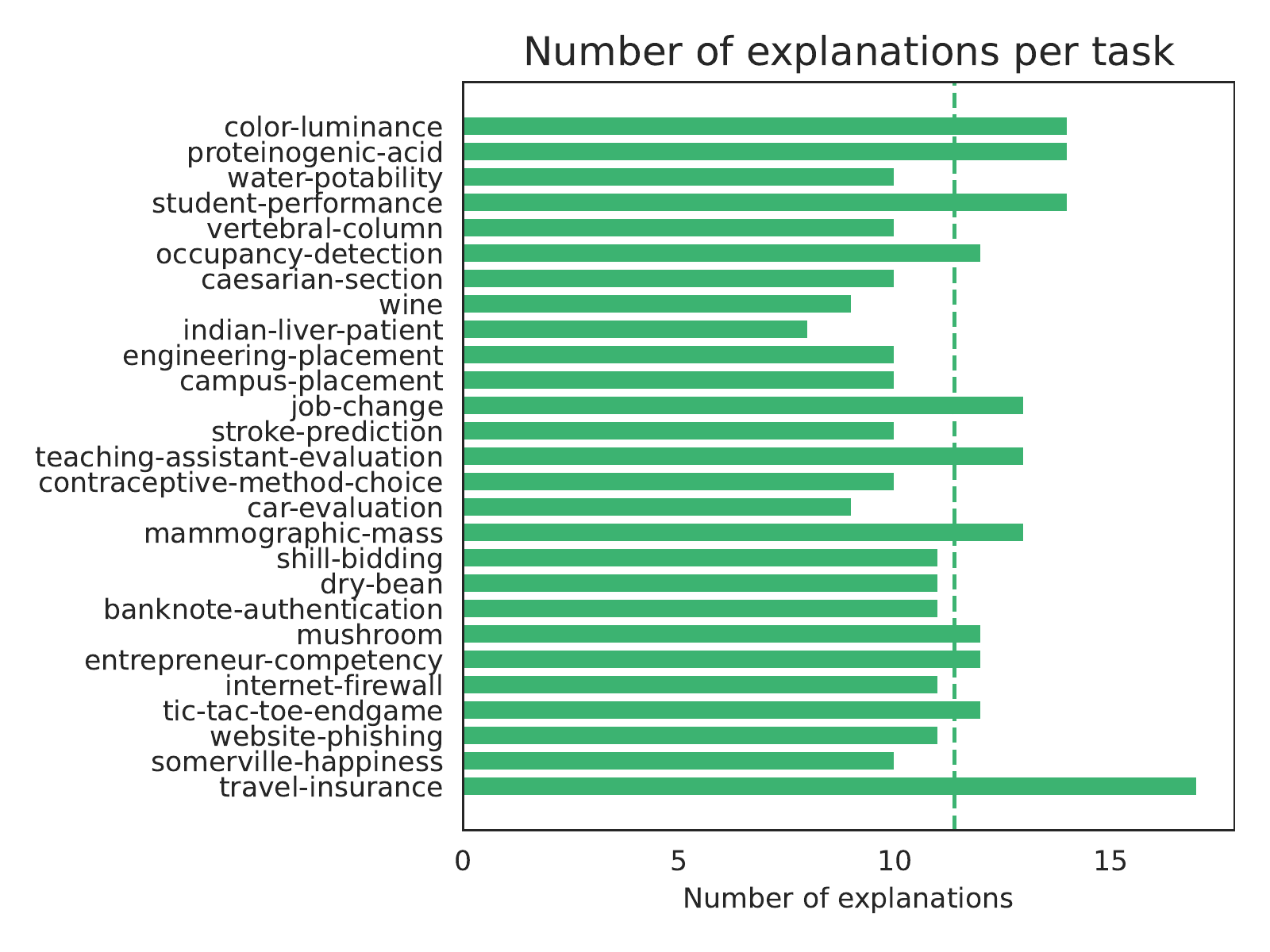}
      \centering
        \begingroup\renewcommand{\caption}[1]{(a)}%
      \caption{Number of Explanations}
      \label{fig:exp_count}
      \endgroup
    \endminipage\hfill
    \minipage{0.33\textwidth}
      \includegraphics[width=\linewidth]{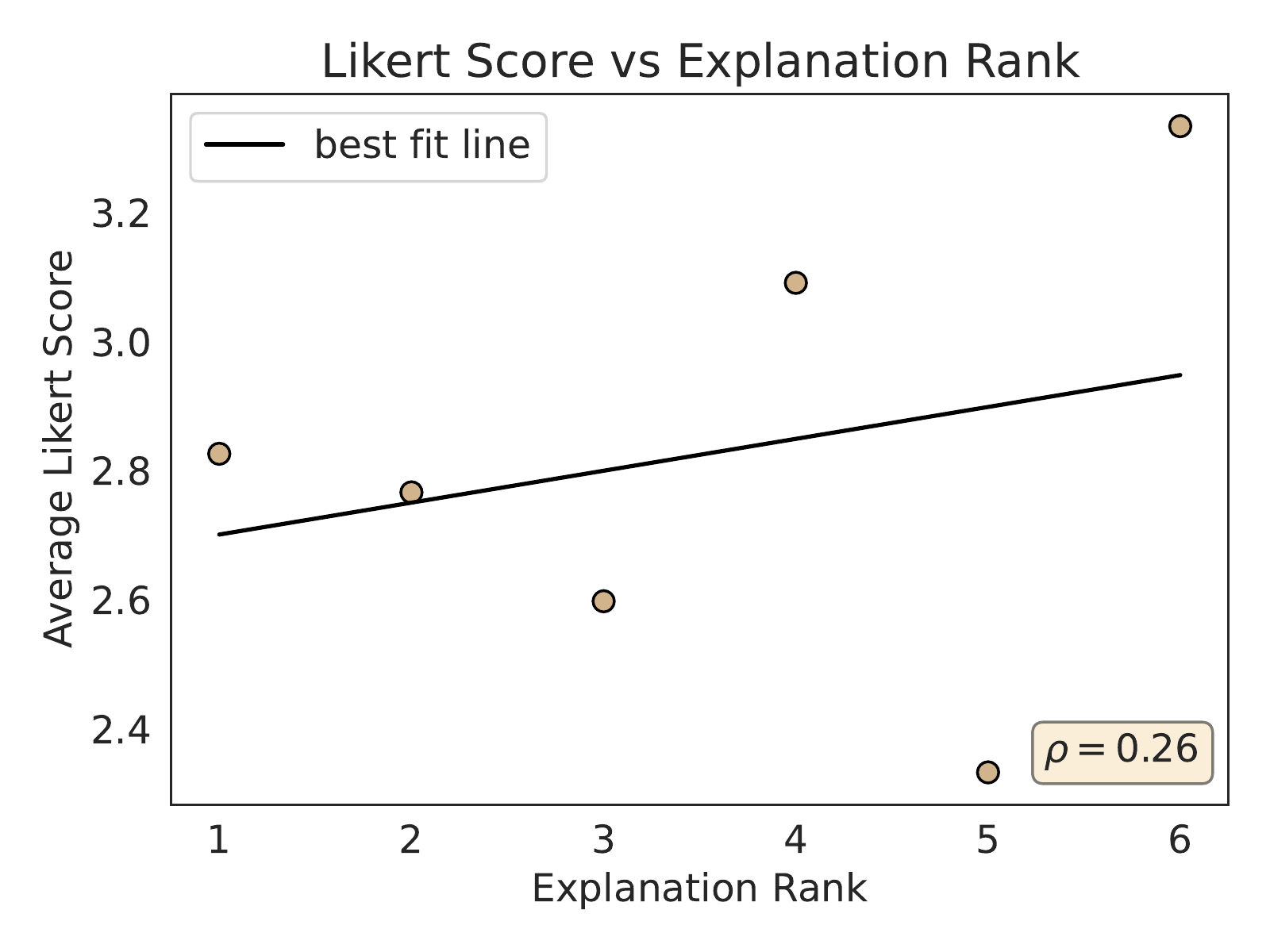}
      \centering
      \begingroup\renewcommand{\caption}[1]{(b)}%
      \caption{Likert Scores vs Rank of Explanations}
      \label{fig:likert_rank}
      \endgroup
    \endminipage\hfill
    \minipage{0.33\textwidth}%
      \includegraphics[width=\linewidth]{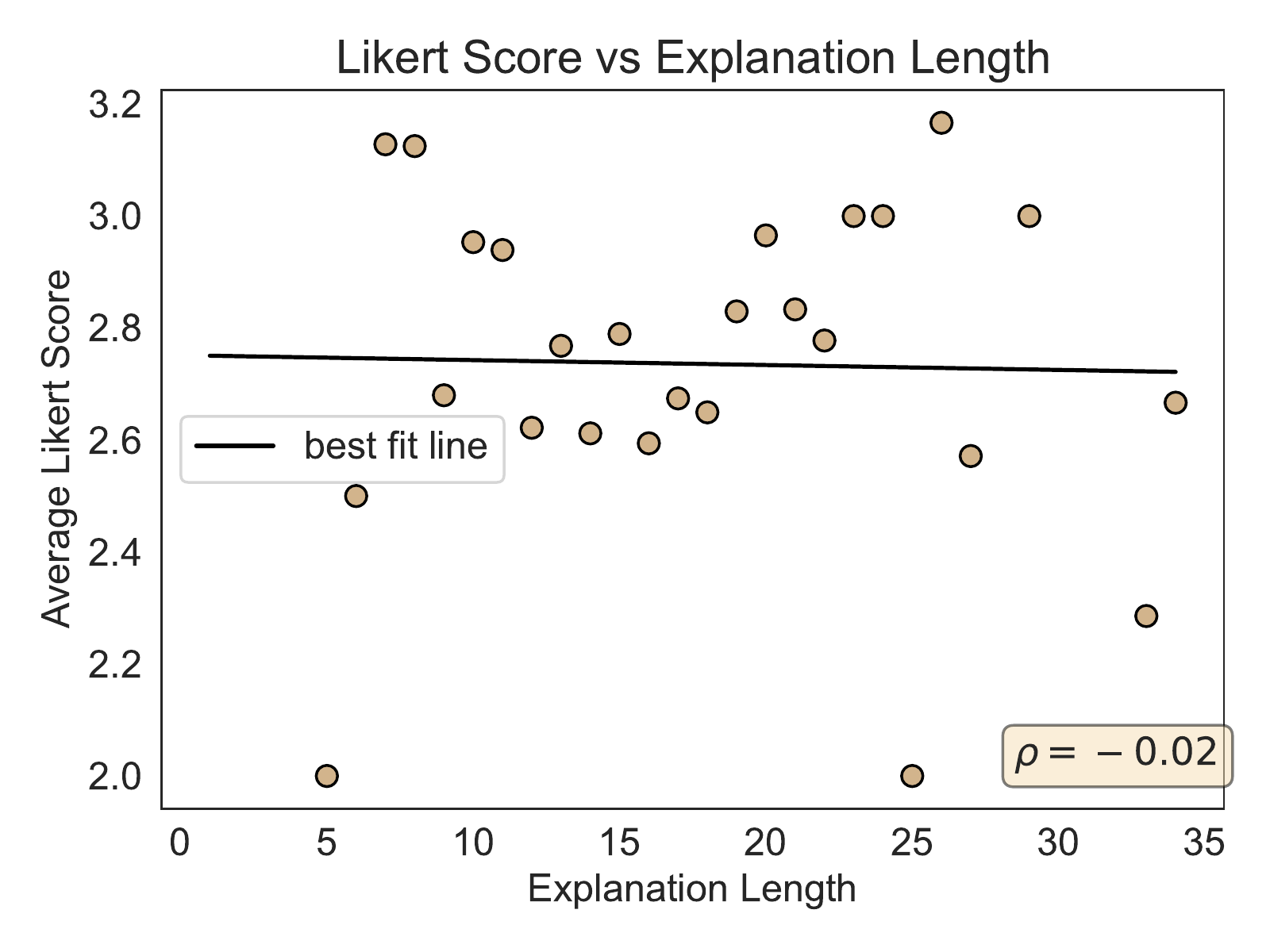}
    \centering
      \begingroup\renewcommand{\caption}[1]{(c)}%
      \caption{Likert Scores vs Explanation Length}
      \label{fig:likert_length}
      \endgroup
    \endminipage
        \caption{(a) On Average we obtain over 10 explanations per task in \benchmarkreal for tasks that are crowdsourced (b) Weak positive correlation indicating later explanations were given higher likert scores by students. Likert ratings were averaged for each rank. (c) Near-zero correlation indicating that likert ratings given by students were almost independent of explanation length. Likert ratings were averaged for each length. ($\rho$ denotes Pearson correlation coefficient in each of the plots)}
    \label{fig:appendixvariations_across_tasks}
\end{figure*}

For the crowdsourced datasets, we show the number of explanations collected per task in Figure~\ref{fig:appendixvariations_across_tasks}(a). The number of explanations is largely around an average value of ~11 explanations per task.

Figure~\ref{fig:appendixvariations_across_tasks}(b) shows the relation between explanation quality (quantified by likert scores) and rank of the explanation. Rank denotes the order in which a teacher provided that explanation during our crowdsourced explanation collection phase. We find a positive correlation between quality and rank of explanation showing that teachers generally submit most useful explanations (as perceived by them) to teach a task. Finally, we do not observe any correlation between explanation length and ratings as indicated by Figure \ref{fig:appendixvariations_across_tasks}(c).

\begin{figure}[!ht]
    \includegraphics[scale=0.47]{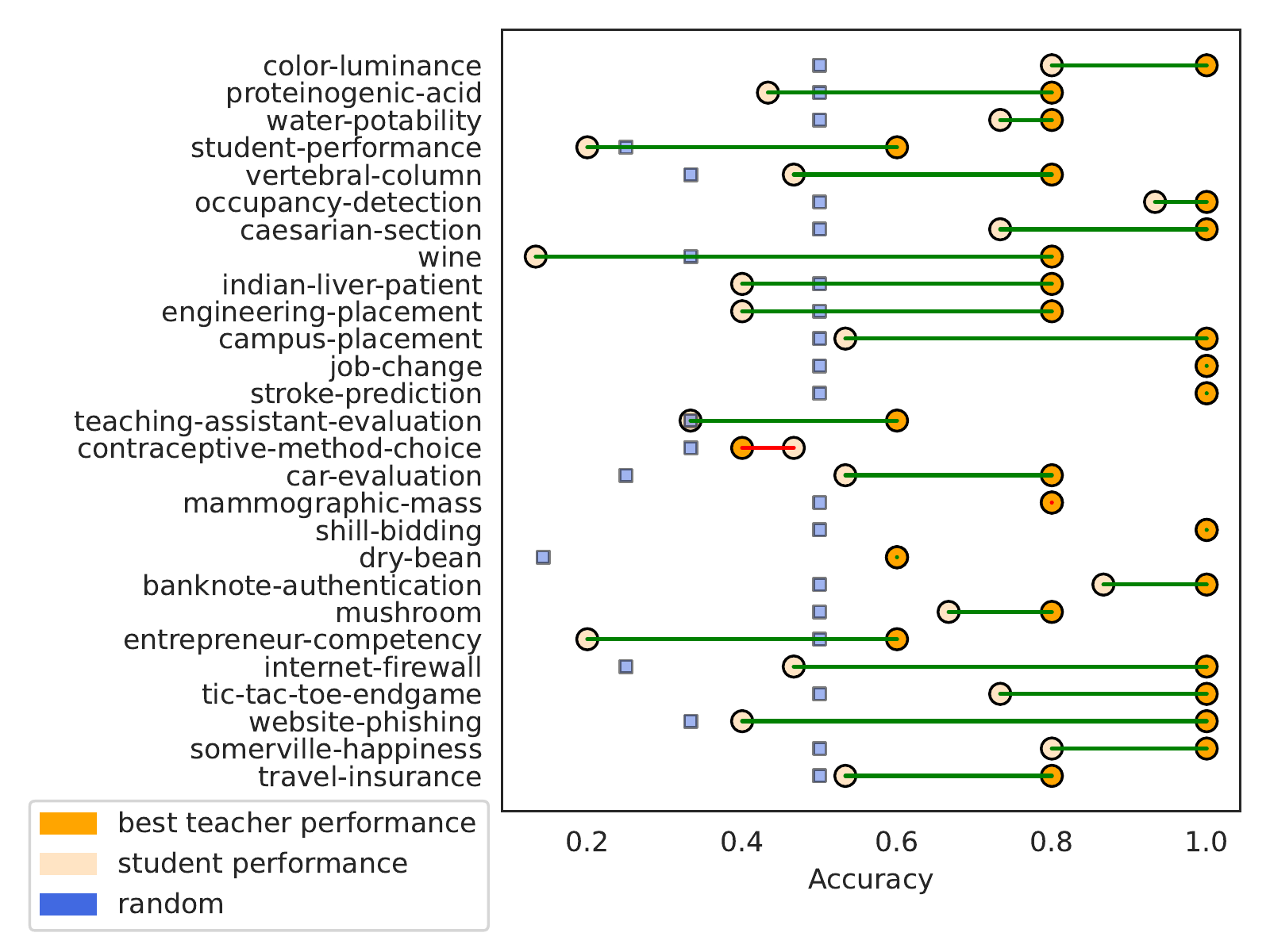}
    \caption{Best teacher vs average of their students for tasks in \benchmarkreal. \textcolor{red}{Red} lines indicate cases where the student performance is more than the teacher performance. {\color{ForestGreen} Green} lines indicate cases where teachers perform better than students.}
    \label{fig:best_teacher_student}
\end{figure}

\begin{figure}[!ht]
    \includegraphics[scale=0.47]{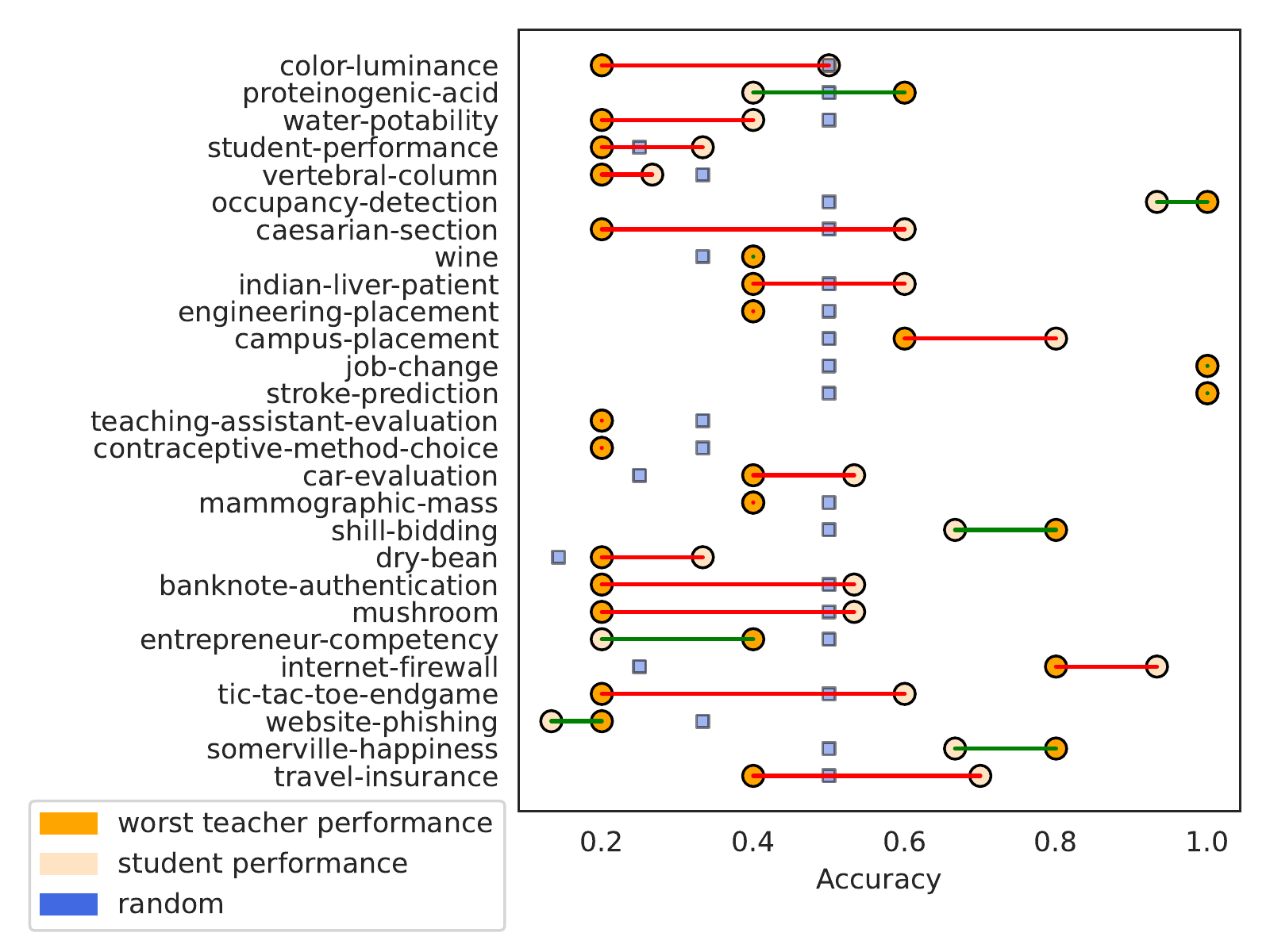}
    \caption{Worst teacher vs average of their students for tasks in \benchmarkreal. \textcolor{red}{Red} lines indicate cases where the student performance is more than the teacher performance. {\color{ForestGreen} Green} lines indicate cases where teachers perform better than students.}
    \label{fig:worst_teacher_student}
\end{figure}

We also illustrate the differences between teacher and student on our tasks in \secref{sec:dataset_analysis}. Here we present two additional plots showing the performance of (1) best teacher vs their students for each task (Figure~\ref{fig:best_teacher_student}) and (2) worst teacher vs their students for each task (Figure~\ref{fig:worst_teacher_student}). We find that even though the best teachers often attain near-perfect accuracies for the tasks, their students perform significantly worse than them in many tasks. The explanations from the worst teachers did not help students in getting significantly better than random performance for majority of the tasks, even though the student did outperform the worst teacher.

\section{Reward Structure for Crowd-sourcing Tasks}\label{sec:appendix_reward}

Our work involves multiple stages of crowdsourcing to collect high-quality explanations for the classification tasks. We pick turkers in the US for explanation collection and verification  tasks (US,UK,NZ, and GB for Wikipedia mining Task) with a 98\% HIT approval rate and a minimum of 1000 HITs approved. In Table \ref{tab:reward_structure}, we summarize the payment structure provided to the turkers on the AMT platform for each of the stages (described in detail in \secref{sec:dataset_collection}) -- (1) Wikipedia mining on tables scraped from Wikipedia, (2) Explanation collection for tables obtained from UCI, Kaggle and Wikipedia, and (3) Explanation validation for collected explanations.
For all the three crowdsourcing tasks, the turkers were compensated fairly and the payment per task is equivalent to an hourly compensation that is greater than minimum wage (based on the median time taken by turkers).

\begin{table}[!ht]
    \centering
    \small
    \begin{tabular}{c|c|c}
        \toprule
        \Thead{Stage} & \Thead{\$/HIT} & \Thead{Bonus}\\
        \midrule
        \makecell{Wikipedia Mining}  & \$3 & \$3-\$4 \tablefootnote{\textcent{50} per table submitted}\\
        \makecell{Explanation Collection} & \$5.5 & -\\
        \makecell{Explanation Validation} &  \$1.2 &  -\\
        \bottomrule
    \end{tabular}
    \caption{Payment structure for AMT Tasks}
    \label{tab:reward_structure}
\end{table}


\section{Training details} \label{sec:model_appendix}
In this section we proved details about implementation of various models, hyperparameter details, and details about hardware and software used along with an estimate of time taken to train the models.
Code and dataset for our paper will be made public upon first publication.

\subsection{Details of seen and novel tasks for \benchmarkreal and \benchmarksyn} \label{sec:task_split}
For \benchmarkreal, we chose the tasks from Wikipedia that have very examples to be part of novel task set. Among the tasks from Kaggle and UCI, we kept tasks with higher number of samples as part of seen tasks (training tasks).
Seen tasks (20) for \benchmarkreal are: 
\begin{itemize}[noitemsep, topsep=0pt, leftmargin=*]
    \item \texttt{website-phishing}
    \item \texttt{internet-firewall} 
    \item \texttt{mushroom}
    \item \texttt{dry-bean} 
    \item \texttt{wine}
    \item \texttt{caesarian-section}
    \item \texttt{occupancy-detection}
    \item \texttt{vertebral-column}
    \item \texttt{student-performance}
    \item \texttt{shill-bidding}
    \item \texttt{mammographic-mass}
    \item \texttt{teaching-assistant-evaluation}
    \item \texttt{somerville-happiness}
    \item \texttt{stroke-prediction}
    \item \texttt{job-change}
    \item \texttt{campus-placement}
    \item \texttt{engineering-placement}
    \item \texttt{water-potability}
    \item \texttt{color-luminance}
    \item \texttt{proteinogenic-acid}
\end{itemize}
\noindent
Novel tasks (16) for \benchmarkreal are:
\begin{itemize}[noitemsep, topsep=0pt, leftmargin=*]
    \item \texttt{banknote-authentication}
    \item \texttt{tic-tac-toe-endgame}
    \item \texttt{car-evaluation}
    \item \texttt{contraceptive-method-choice}
    \item \texttt{indian-liver-patient}
    \item \texttt{travel-insurance}
    \item \texttt{entrepreneur-competency}
    \item \texttt{award-nomination-result}
    \item \texttt{coin-face-value}
    \item \texttt{coin-metal}
    \item \texttt{driving-championship-points}
    \item \texttt{election-outcome}
    \item \texttt{hotel-rating}
    \item \texttt{manifold-orientability}
    \item \texttt{soccer-club-region}
    \item \texttt{soccer-league-type}
\end{itemize}

We train on 70\% of the labeled examples of the seen tasks and perform zero-shot generalization test over the 20\% examples of each task in \benchmarkreal. For the extremely small Wikipedia tasks (for which we do not crowdsource explanations), we use all examples for zero-shot testing.

For \benchmarksyn, we have 96 tasks as seen (training) tasks and 48 as novel tasks. Task in \benchmarksyn that belong to the following schemas are part of the seen tasks:
\begin{itemize}[noitemsep, topsep=0pt, leftmargin=*]
    \item Species of Animal
    \item Species of Bird
    \item Rainfall prediction
\end{itemize}
Tasks belonging to `Bond relevance classification' and `League Rank Classification' were part of novel tasks for \benchmarksyn.
We train on 700 labeled examples of each seen task and perform zero-shot generalization test over 200 examples of each novel task in \benchmarksyn.

\subsection{Model parameters}
\begin{itemize}[noitemsep, topsep=0pt, leftmargin=*]
    \item \underline{RoBERTa w/o Exp.}: The number of parameters is same as the pretrained RoBERTa-base model available on HuggingFace library. 
    \item \underline{RoBERTa w/ Exp.}: The number of parameters is same as the pretrained RoBERTa-base model available on HuggingFace library. 
    \item \underline{\model}: The number of parameters is same as the pre-trained RoBERTa mdoel finetuned on MNLI~\cite{williams2017broad} corpus. We obtain the pretrained checkpoint from HuggingFace.\footnote{Weights link: \url{https://huggingface.co/textattack/roberta-base-MNLI}}
\end{itemize}

\subsection{Hyper-parameter settings}
For all the transformer based models we use the implementation of HuggingFace library \cite{wolf-etal-2020-transformers}. All the model based hyper-parameters are thus kept default to the settings in the HuggingFace library. We use the publicly available checkpoints to initialise the pre-trained models.
For RoBERTa based baselines we use `roberta-base' checkpoint available on HuggingFace.
For our intermediate entailment model in \model, we finetune a pretrained checkpoint of RoBERTa trained on MNLI corpus (`textattack/roberta-base-MNLI') 

When training on \benchmarksyn, we use a maximum of 64 tokens for our baseline RoBERTa w/o Exp. and \model. For the RoBERTa w/ Exp. model we increase this limit to 128 tokens as it takes concatenation of all explanations for a task.
When training on \benchmarkreal, we use 256 tokens as limit for RoBERTa w/ Exp. using explanations as the real-world tasks have roughly two times more explanations on average than synthetic tasks.

We used the AdamW \cite{loshchilov2018decoupled} optimizer commonly used to fine-tune pre-trained Masked Language Models (MLM) models. 
For fine-tuning the pre-trained models on our benchmark tasks, we experimented with learning rates \{$1e-5, 2e-5$\} and chose $1e-5$ based on performance on the performance on the validation set of seen tasks.
Batch sizes was kept as 2 with gradient accumulation factor of 8. The random seed for all experiments was 42. 
We train all the models for 20 epochs. Each epoch comprises of 100 batches, and in each batch the models look at one of the tasks in the seen split.

\subsection{Hardware and software specifications}
All the models are coded using Pytorch 1.4.0\footnote{\url{https://pytorch.org/}} \cite{NEURIPS2019_9015} and related libraries like numpy \cite{harris2020array}, scipy \cite{jones2001scipy} etc. 
We run all experiments on one of the following two systems - (1) GeForce RTX 2080 GPU of size 12 GB, 256 GB RAM and 40 CPU cores (2) Tesla V100-SXM2 GPU of size 16GB, 250 GB RAM and 40 CPU cores. 
\subsection{Training times}
\begin{itemize}[noitemsep, topsep=0pt, leftmargin=*]
    \item Training on \benchmarkreal: The baseline RoBERTa w/o Exp model typically takes 3 seconds on average for training on 1 batch of examples. In 1 batch, the model goes through 16 examples from the tasks in seen split. 
    RoBERTa w/ Exp. takes around 5 seconds to train on 1 batch. \model takes longer time than baselines owing to the multiple forward passes. For training on 1 batch of \benchmarkreal, \model took 12 seconds on average.
    \item Training on \benchmarksyn: All the models take comparatively much lesser time for training on our synthetic tasks owing to lesser number of explanations on average for a task. For training on 1 batch, all models took 1 seconds or less to train on 1 batch of examples from \benchmarksyn.
\end{itemize}

\section{Effect of scrambling attribute names in input to \model}
We performed an additional experiment on our synthetic data to evaluate if \model can (1) understand the relationship between attribute names and attribute values and (2) identify the correspondence between attribute names in the explanations with the attribute name-value pair in the FaT representation of the structured input.
In this experiment, we scrambled (randomly permuted the column names) the structured input when performing inference over the tasks in \benchmarksyn. 
So after the scrambling operation, the attribute name-value pairs will be incoherent.
We considered 5 random seeds (42 to 46) for this scrambling operation.
The mean generalization performance (accuracy) using scrambled inputs for \benchmarksyn is 41.62\% (with standard deviation as 0.9\%).
This is comparable with the random baseline on \benchmarksyn (42.19\%) as expectedly, \model fails to identify meaningful correspondences between the explanation and input when the inputs are incoherent.

\section{Annotation interfaces} \label{sec:annotation_interface}
\begin{figure*}[!ht]
    \centering
    \includegraphics[scale=0.4]{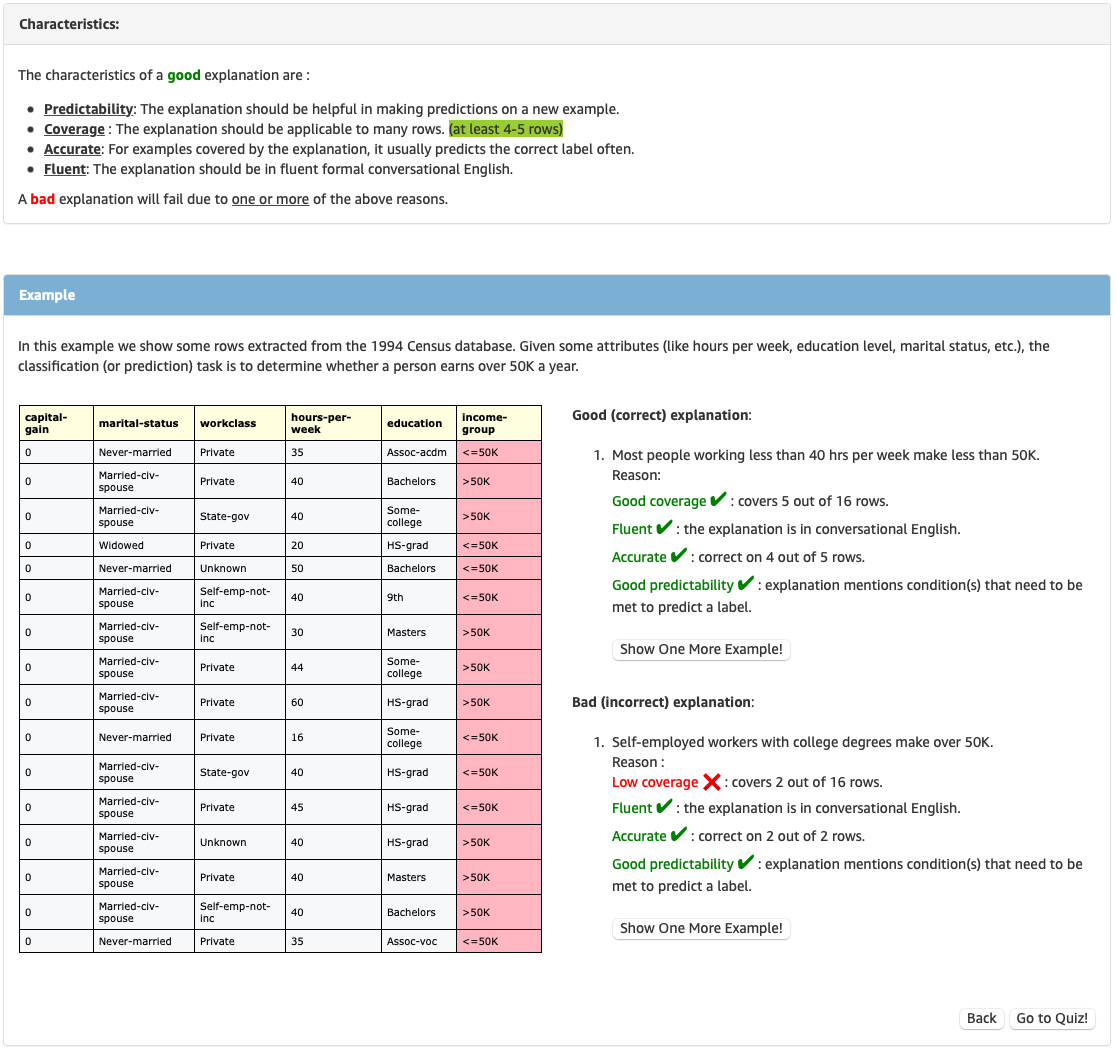}
    \caption{Explanation Collection: Annotation Task Examples page.}
    \label{fig:aninf_teacher_1}
\end{figure*}

\begin{figure*}
    \centering
    \includegraphics[scale=0.4]{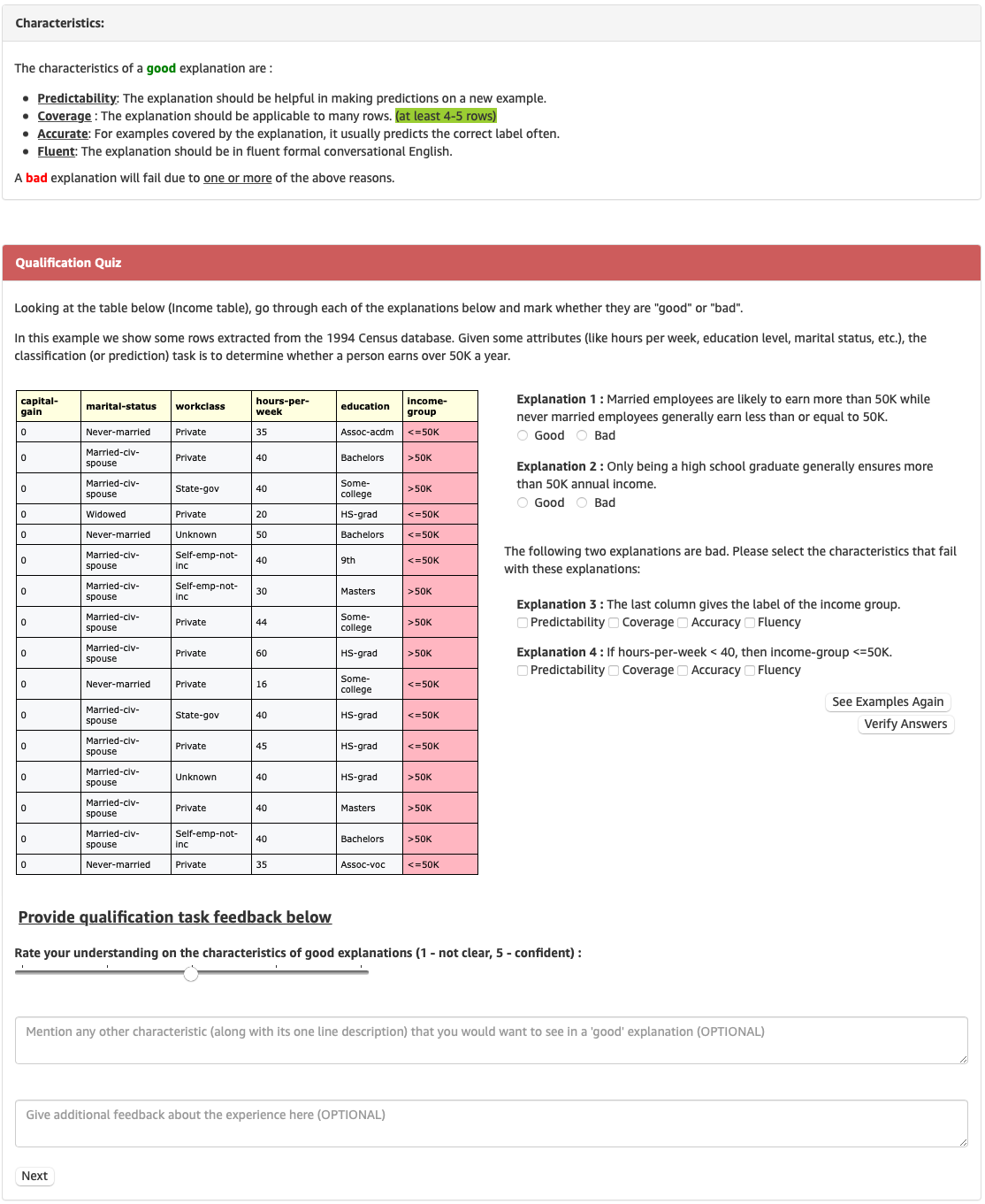}
    \caption{Explanation Collection: Qualification Task page.}
    \label{fig:aninf_teacher_2}
\end{figure*}

\begin{figure*}
    \centering
    \includegraphics[scale=0.4]{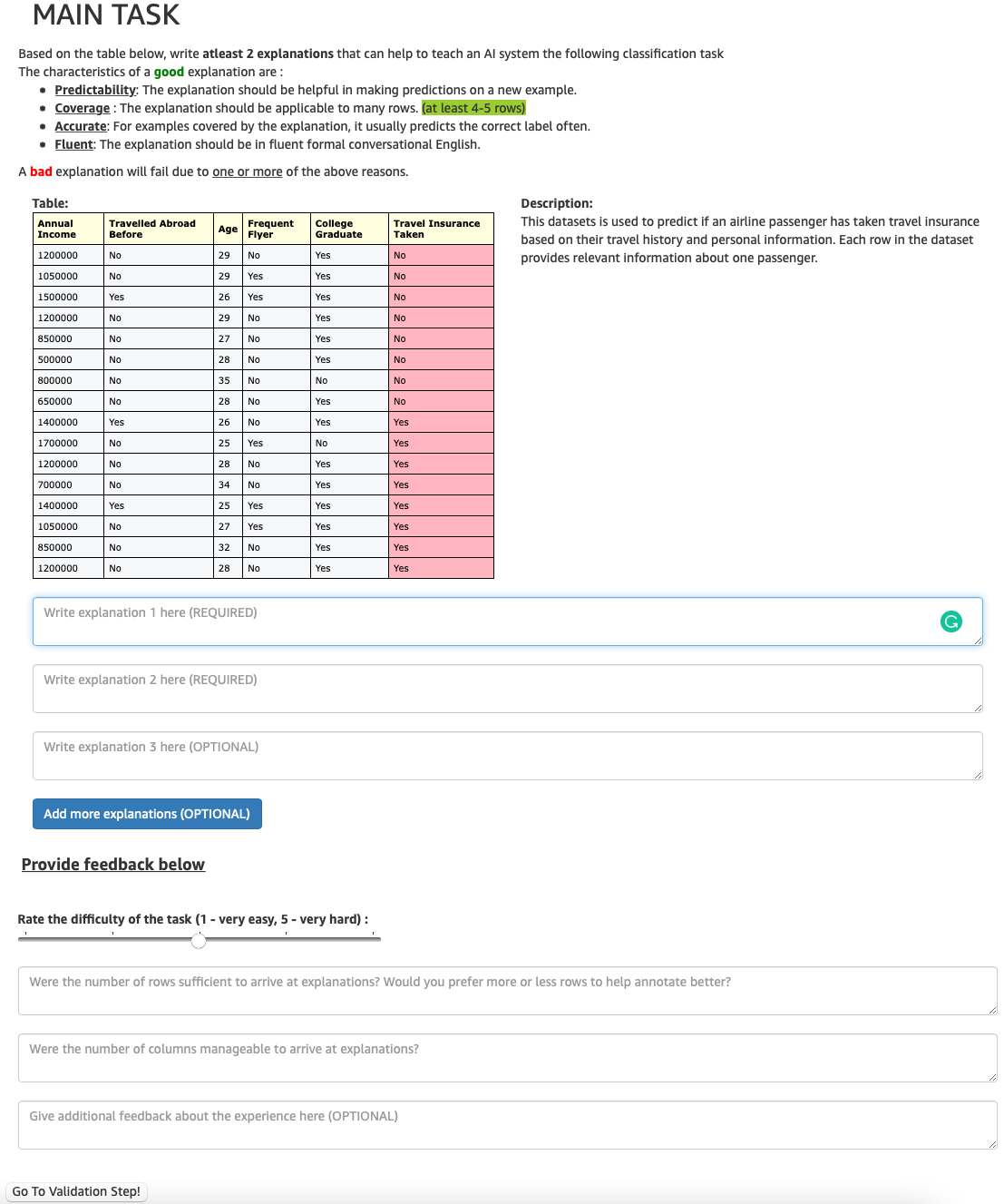}
    \caption{Explanation Collection: Main Task page.}
    \label{fig:aninf_teacher_3}
\end{figure*}

\begin{figure*}
    \centering
    \includegraphics[scale=0.4]{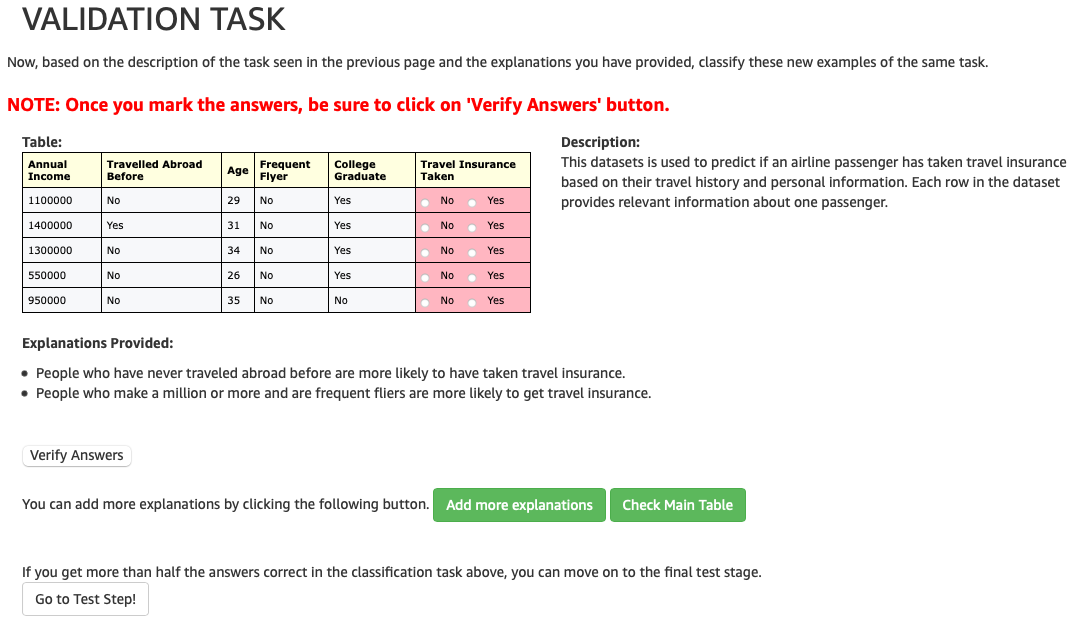}
    \includegraphics[scale=0.4]{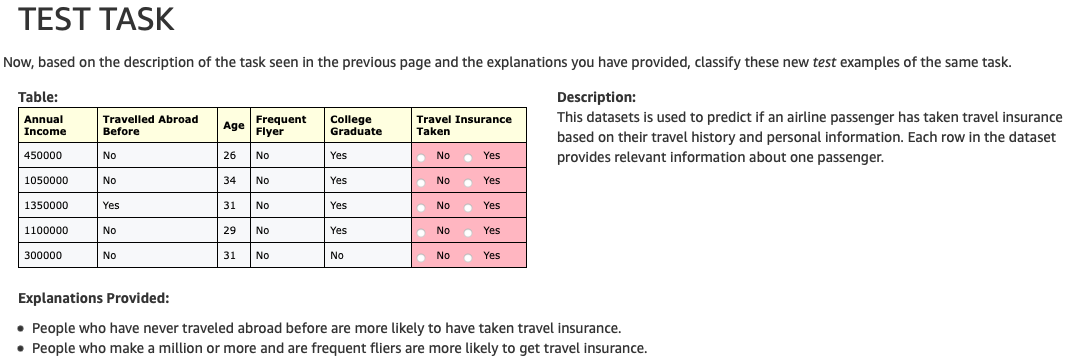}
    \caption{Explanation Collection: Validation and Test page.}
    \label{fig:aninf_teacher_4}
\end{figure*}

\begin{figure*}
    \centering
    \includegraphics[scale=0.39]{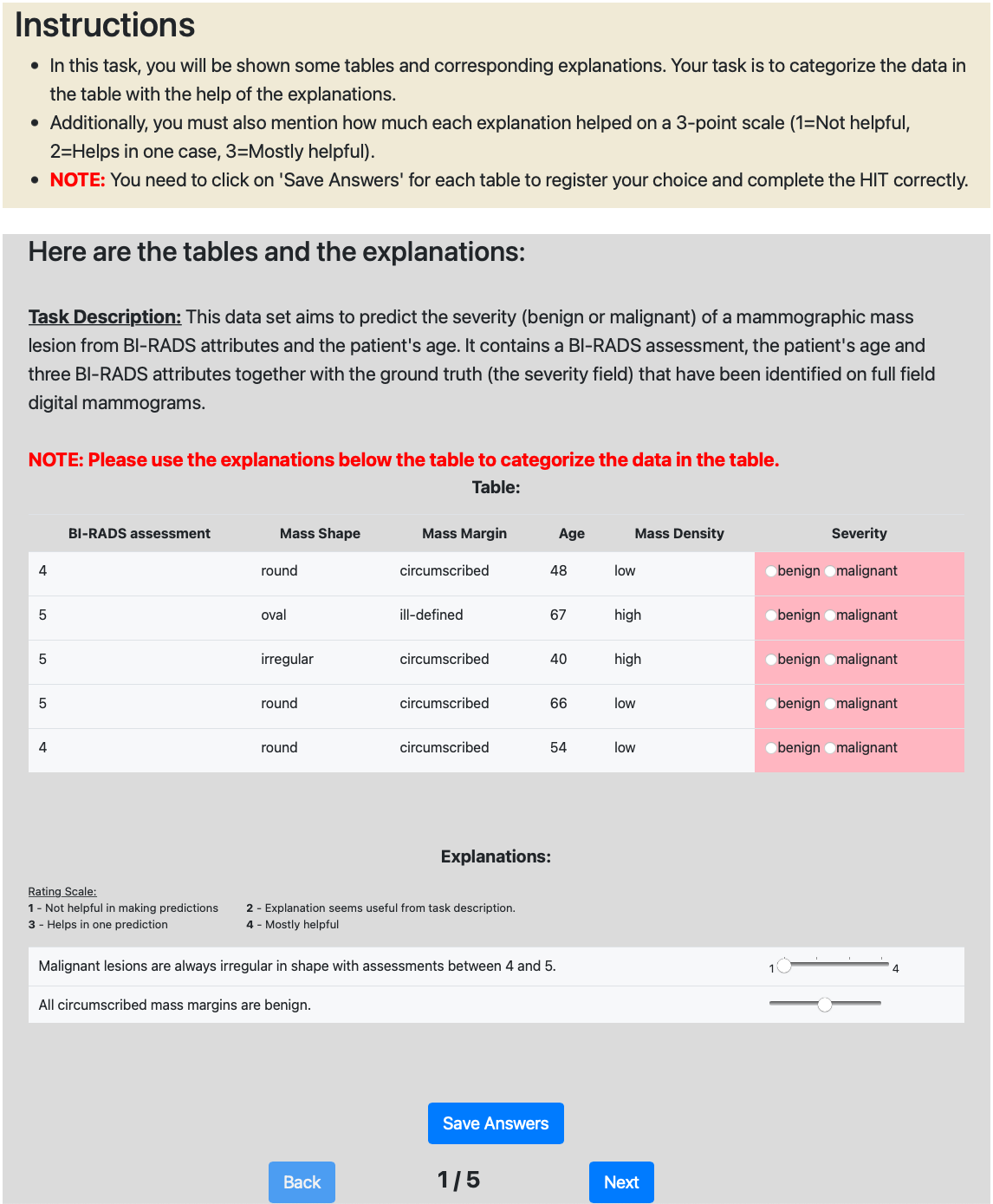}
    \caption{Explanation Verification page.}
    \label{fig:aninf_student}
\end{figure*}

We present the different annotation templates and interfaces used for our explanation collection and verification stages in Figures \ref{fig:aninf_teacher_1},\ref{fig:aninf_teacher_2},\ref{fig:aninf_teacher_3},\ref{fig:aninf_teacher_4} and Figure \ref{fig:aninf_student} respectively. 
\pagebreak
\begin{table*}[t!]
\resizebox{\textwidth}{!}{
\begin{tabular}{l|l|l|c}
\hline
\Thead{Dataset}                       & \Thead{Source} & \Thead{URL}  & \Thead{Crowd-sourced}\\
\midrule
\texttt{car-evaluation}                & UCI    & {\footnotesize \url{ https://archive.ics.uci.edu/ml/datasets/Car+Evaluation}}   & YES                                        \\
\texttt{indian-liver-patient}          & UCI    & {\footnotesize \url{ https://archive.ics.uci.edu/ml/datasets/ILPD+\%28Indian+Liver+Patient+Dataset\%29}}   & YES                  \\
\texttt{bank-note-authentication}      & UCI    & {\footnotesize \url{ http://archive.ics.uci.edu/ml/datasets/banknote+authentication}}      & YES                                  \\
\texttt{contraceptive-method-choice}   & UCI    & {\footnotesize \url{ http://archive.ics.uci.edu/ml/datasets/Contraceptive+Method+Choice}} & YES                                   \\
\texttt{mushroom}                      & UCI    & {\footnotesize \url{ http://archive.ics.uci.edu/ml/datasets/Mushroom}}                      & YES                                 \\
\texttt{mammographic-mass}             & UCI    & {\footnotesize \url{ https://archive.ics.uci.edu/ml/datasets/Mammographic+Mass}}        & YES                                     \\
\texttt{wine}                          & UCI    & {\footnotesize \url{ http://archive.ics.uci.edu/ml/datasets/Wine}}                         & YES                                  \\
\texttt{teaching-assistant-evaluation} & UCI    & {\footnotesize \url{ https://archive.ics.uci.edu/ml/datasets/Teaching+Assistant+Evaluation}}  & YES                               \\
\texttt{shill-bidding}                 & UCI    & {\footnotesize \url{ https://archive.ics.uci.edu/ml/datasets/Shill+Bidding+Dataset}}        & YES                                 \\
\texttt{website-phishing}              & UCI    & {\footnotesize \url{ https://archive.ics.uci.edu/ml/datasets/Website+Phishing}}             & YES                                 \\
\texttt{tic-tac-toe-endgame}           & UCI    & {\footnotesize \url{ https://archive.ics.uci.edu/ml/datasets/Tic-Tac-Toe+Endgame}}           & YES                                \\
\texttt{somerville-happiness}          & UCI    & {\footnotesize \url{ https://archive.ics.uci.edu/ml/datasets/Somerville+Happiness+Survey}}   & YES                                \\
\texttt{occupancy-detection}           & UCI    & {\footnotesize \url{ https://archive.ics.uci.edu/ml/datasets/Occupancy+Detection+}}          & YES                                \\
\texttt{vertebral-column}              & UCI    & {\footnotesize \url{ https://archive.ics.uci.edu/ml/datasets/Vertebral+Column}}             & YES                                 \\
\texttt{caesarian-section}             & UCI    & {\footnotesize \url{ https://archive.ics.uci.edu/ml/datasets/Caesarian+Section+Classification+Dataset}}     & YES                 \\
\texttt{student-performance}           & UCI    & {\footnotesize \url{ https://archive.ics.uci.edu/ml/datasets/Student+Performance+on+an+entrance+examination}}   & YES             \\
\texttt{dry-bean}                      & UCI    & {\footnotesize \url{ https://archive.ics.uci.edu/ml/datasets/Dry+Bean+Dataset}}             & YES                                 \\
\texttt{internet-firewall}             & UCI    & {\footnotesize \url{ https://archive.ics.uci.edu/ml/datasets/Internet+Firewall+Data}}         & YES                               \\
\texttt{campus-placement}              & Kaggle & {\footnotesize \url{ https://www.kaggle.com/benroshan/factors-affecting-campus-placement}}  & YES                                 \\
\texttt{job-change}                    & Kaggle & {\footnotesize \url{ https://www.kaggle.com/arashnic/hr-analytics-job-change-of-data-scientists?select=aug\_train.csv}} & YES     \\
\texttt{water-potability}              & Kaggle & {\footnotesize \url{ https://www.kaggle.com/adityakadiwal/water-potability}}   & YES                                              \\
\texttt{stroke-prediction}             & Kaggle & {\footnotesize \url{ https://www.kaggle.com/fedesoriano/stroke-prediction-dataset}}     & YES                                     \\
\texttt{engineering-placement}         & Kaggle & {\footnotesize \url{ https://www.kaggle.com/tejashvi14/engineering-placements-prediction}}     & YES                              \\
\texttt{travel-insurance}              & Kaggle & {\footnotesize \url{ https://www.kaggle.com/tejashvi14/travel-insurance-prediction-data}}     & YES                               \\
\texttt{entrepreneur-competency}       & Kaggle & {\footnotesize \url{ https://www.kaggle.com/namanmanchanda/entrepreneurial-competency-in-university-students}} & YES     \\
\texttt{soccer-league-type}       & Wikipedia & {\footnotesize \url{https://en.wikipedia.org/wiki/Oklahoma}} & NO     \\
\texttt{soccer-club-region}       & Wikipedia & {\footnotesize \url{https://en.wikipedia.org/wiki/Oklahoma}} & NO     \\
\texttt{hotel-rating}       & Wikipedia & {\footnotesize \url{https://en.wikipedia.org/wiki/Disneyland_Paris}} & NO     \\
\texttt{coin-face-value}       & Wikipedia & {\footnotesize \url{https://en.wikipedia.org/wiki/Coins_of_the_United_States_dollar}} & NO     \\
\texttt{coin-metal}       & Wikipedia & {\footnotesize \url{https://en.wikipedia.org/wiki/Coins_of_the_United_States_dollar}} & NO     \\
\texttt{election-outcome}       & Wikipedia & {\footnotesize \url{https://en.wikipedia.org/wiki/Kuomintang}} & NO     \\
\texttt{driving-championship-points}       & Wikipedia & {\footnotesize \url{https://en.wikipedia.org/wiki/Judd_(engine)}} & NO     \\
\texttt{manifold-orientability}       & Wikipedia & {\footnotesize \url{https://en.wikipedia.org/wiki/Homology_(mathematics)}} & NO     \\
\texttt{award-nomination-result}       & Wikipedia & {\footnotesize \url{https://en.wikipedia.org/wiki/When_Harry_Met_Sally...}} & NO     \\
\texttt{color-luminance}       & Wikipedia & {\footnotesize \url{https://en.wikipedia.org/wiki/Hue}} & YES     \\
\texttt{proteinogenic-acid}       & Wikipedia & {\footnotesize \url{https://en.wikipedia.org/wiki/Miller\%E2\%80\%93Urey_experiment}} & YES     \\
\hline
\end{tabular}}
\caption{List of datasets and URLs that make up \benchmarkreal.}
\label{tab:benchmark_source}
\end{table*}

\colorlet{punct}{red!60!black}
\definecolor{background}{HTML}{EEEEEE}
\definecolor{delim}{RGB}{20,105,176}
\colorlet{numb}{magenta!60!black}

\lstdefinelanguage{json}{
    basicstyle=\normalfont\ttfamily,
    numbers=left,
    numberstyle=\scriptsize,
    stepnumber=1,
    numbersep=8pt,
    showstringspaces=false,
    breaklines=true,
    frame=lines,
    backgroundcolor=\color{background},
    literate=
     *{0}{{{\color{numb}0}}}{1}
      {1}{{{\color{numb}1}}}{1}
      {2}{{{\color{numb}2}}}{1}
      {3}{{{\color{numb}3}}}{1}
      {4}{{{\color{numb}4}}}{1}
      {5}{{{\color{numb}5}}}{1}
      {6}{{{\color{numb}6}}}{1}
      {7}{{{\color{numb}7}}}{1}
      {8}{{{\color{numb}8}}}{1}
      {9}{{{\color{numb}9}}}{1}
      {:}{{{\color{punct}{:}}}}{1}
      {,}{{{\color{punct}{,}}}}{1}
      {\{}{{{\color{delim}{\{}}}}{1}
      {\}}{{{\color{delim}{\}}}}}{1}
      {[}{{{\color{delim}{[}}}}{1}
      {]}{{{\color{delim}{]}}}}{1},
}

\begin{table*}
\small
\begin{lstlisting}[language=json,numbers=none]
    "description": "This dataset is used to predict the type of birds based on the given attributes. Each row provides the relevant attributes of a bird.",
        "column_names":{
            "size" : ["categorical", ["large", "medium", "small"]],
            "size (number)" : ["number", [10, 100]],
            "color" : ["categorical", ["red", "blue", "green", "brown", "pink", "orange", "black", "white"]],
            "head" : ["categorical", ["yes", "no"]],
            "length" : ["categorical", ["tall", "medium", "short"]],
            "length (number)" : ["number", [10,100]],
            "tail" : ["categorical", ["yes", "no"]],
            "number of faces" : ["number", [1,3]],
            "arms" : ["categorical", ["yes", "no"]],
            "legs" : ["categorical", [2, 4, 6, 8]],
            "hair" : ["categorical", ["yes", "no"]],
            "wings" : ["categorical", ["yes", "no"]],
            "feathers" : ["categorical", ["yes", "no"]],
            "airborne" : ["categorical", ["yes", "no"]],
            "toothed" : ["categorical", ["yes", "no"]],
            "backbone" : ["categorical", ["yes", "no"]],
            "venomous" : ["categorical", ["yes", "no"]],
            "domestic" : ["categorical", ["yes", "no"]],
            "region": ["categorical", ["asia", "europe", "americas", "africas", "antartica", "oceania"]]
        },
        "targets": {
            "bird species": ["wug", "blicket", "dax", "toma", "pimwit", "zav", "speff", "tulver", "gazzer", "fem", "fendle", "tupa"]
        } 
}
\end{lstlisting}
\caption{Synthetic table schema 1: Species of Birds}
\label{tab:schema_birds}
\end{table*}

\begin{table*}
\small
\begin{lstlisting}[language=json,numbers=none]
{
    "description": "This dataset is used to predict the type of an aquatic animal based on the given attributes. Each row provides the relevant attributes of an animal.",
        "column_names":{
            "size" : ["categorical", ["large", "medium", "small"]],
            "size (number)" : ["number", [10, 100]],
            "color" : ["categorical", ["red", "blue", "green", "brown", "pink", "orange", "black", "white"]],
            "head" : ["categorical", ["yes", "no"]],
            "length" : ["categorical", ["tall", "medium", "short"]],
            "length (number)" : ["number", [10,100]],
            "tail" : ["categorical", ["yes", "no"]],
            "number of faces" : ["number", [1,3]],
            "arms" : ["categorical", ["yes", "no"]],
            "legs" : ["categorical", ["yes", "no"]],
            "hair" : ["categorical", ["yes", "no"]],
            "fins" : ["categorical", ["yes", "no"]],
            "toothed" : ["categorical", ["yes", "no"]],
            "venomous" : ["categorical", ["yes", "no"]],
            "domestic" : ["categorical", ["yes", "no"]],
            "region": ["categorical", ["atlantic", "pacific", "indian", "arctic"]]
        },
        "targets": {
            "animal species": ["wug", "blicket", "dax", "toma", "pimwit", "zav", "speff", "tulver", "gazzer", "fem", "fendle", "tupa"]
        } 
}
\end{lstlisting}
\caption{Synthetic table schema 2: Species of Animal}
\label{tab:schema_animals}
\end{table*}

\begin{table*}
\small
\begin{lstlisting}[language=json,numbers=none]
{
   "description": "This dataset is used to predict if it will rain tomorrow or not based on the given attributes. Each row provides the relevant attributes of a day.",
        "column_names":{
            "location" : ["categorical", ["sphinx", "doshtown", "kookaberra", "shtick union", "dysyen"]],
            "mintemp": ["number", [1,15]],
            "maxtemp": ["number", [17,35]],
            "rainfall today": ["categorical", [0, 0.2, 0.4, 0.6, 0.8, 1]],
            "hours of sunshine": ["categorical", [0, 4, 8, 12]],
            "humidity": ["number", [0,100]],
            "wind direction": ["categorical", ["N", "S", "E", "W", "NW", "NE", "SE", "SW"]],
            "wind speed": ["number", [10,85]],
            "atmospheric pressure": ["number", [950,1050]]
        },
        "targets": {
            "rain tomorrow": ["yes", "no"]
        }
}
\end{lstlisting}
\caption{Synthetic table schema 3: Rainfall Prediction}
\label{tab:schema_rain}
\end{table*}

\begin{table*}
\small
\begin{lstlisting}[language=json,numbers=none]
{
    "description": "This dataset is used to predict the final league position of a team based on the given attributes. Each row provides the relevant attributes of a team.",
        "column_names":{
            "win percentage":["number", [0,100]],
            "adjusted offensive efficiency": ["number", [0,100]],
            "adjusted defensive efficiency": ["number", [0,100]],
            "power rating": ["categorical", [1,2,3,4,5]],
            "turnover percentage": ["number", [0,100]],
            "field goal rating": ["categorical", [1,2,3,4,5]],
            "free throw rating": ["categorical", [1,2,3,4,5]],
            "two point shoot percentage": ["number", [0,100]],
            "three point shoot percentage": ["number", [0,100]]
        },
        "targets": {
            "final position": ["1", "2", "3", "4", "Not Qualified"]
        }
}
\end{lstlisting}
\caption{Synthetic table schema 4: League Ranking Classification}
\label{tab:schema_team}
\end{table*}

\begin{table*}
\small
\begin{lstlisting}[language=json,numbers=none]
{
   "description": "This dataset is used to predict the relevance (higher the better) of a bond to a user based on the given attributes. Each row provides the relevant attributes of a user.",
        "column_names":{
            "user age":["number", [15,65]],
            "user knowledge": ["categorical", [1,2,3,4,5]],
            "user gender": ["categorical", ["male", "female"]],
            "user loyalty": ["categorical", [1,2,3,4,5]],
            "user income": ["number", [1000,10000]],
            "user marital status": ["categorical", ["yes", "no"]],
            "user dependents": ["number", [0,3]]
        },
        "targets": {
            "relevance score": ["1", "2", "3", "4", "5"]
        }
}
\end{lstlisting}
\caption{Synthetic table schema 5: Bond Relevance Classification}
\label{tab:schema_product}
\end{table*}

\pagebreak

\end{document}